\documentclass[conference]{IEEEtran}
\IEEEoverridecommandlockouts
\usepackage{subcaption}
\usepackage[ruled,linesnumbered]{algorithm2e}
\usepackage{amsmath}
\usepackage{amsfonts}
\usepackage{graphicx}
\usepackage{color}
\usepackage{subcaption}
\usepackage{bm}
\usepackage{cite}
\usepackage{balance}
\usepackage{multirow}
\usepackage{todonotes}
\usepackage{xcolor}

\def\BibTeX{{\rm B\kern-.05em{\sc i\kern-.025em b}\kern-.08em
    T\kern-.1667em\lower.7ex\hbox{E}\kern-.125emX}}

\hfuzz=\maxdimen
\definecolor{ScarletRed}{rgb}{0.80,0.00,0.00}
\definecolor{DeepGreen}{rgb}{0.00, 0.39, 0.00}
\definecolor{Honeydew}{rgb}{0.94, 1, 0.94}

\DeclareMathOperator*{\argmax}{arg\,max}
\DeclareMathOperator*{\argmin}{arg\,min}

\newlength\mylength
\newcommand{\mb}[1]{\mathbf{#1}}

\begin{document}

\title{Voronoi-based Efficient Surrogate-assisted Evolutionary Algorithm for Very Expensive Problems\\
% {\footnotesize \textsuperscript{*}Note: Sub-titles are not captured in Xplore and
% should not be used}
\thanks{The paper has been published in CEC2019 conference. Please cite the conference version. 
Links: https://ieeexplore.ieee.org/document/8789910 \ 
DOI: 10.1109/CEC.2019.8789910 }
}

\author{ 

\IEEEauthorblockN{
    Hao Tong, Changwu Huang, Jialin Liu, and Xin Yao \\
}

\IEEEauthorblockA{{Shenzhen Key Laboratory of Computational Intelligence}\\ 
University Key Laboratory of Evolving Intelligent Systems of Guangdong Province \\
Department of Computer Science and Engineering \\
{Southern University of Science and Technology, Shenzhen 518055, China} \\
}

Emails: htong6@outlook.com, \{huangcw3, liujl, xiny\}@sustech.edu.cn
}

\maketitle

\bstctlcite{IEEEexample:BSTcontrol} 

\begin{abstract}

Very expensive problems are very common in practical system that one fitness evaluation costs several hours or even days. Surrogate assisted evolutionary algorithms (SAEAs) have been widely used to solve this crucial problem in the past decades. However, most studied SAEAs focus on solving problems with a budget of at least ten times of the dimension of problems which is unacceptable in many very expensive real-world problems. In this paper, we employ Voronoi diagram to boost the performance of SAEAs and propose a novel framework named Voronoi-based efficient surrogate assisted evolutionary algorithm (VESAEA) for very expensive problems, in which the optimization budget, in terms of fitness evaluations, is only 5 times of the problem’s dimension. In the proposed framework, the Voronoi diagram divides the whole search space into several subspace and then the local search is operated in some potentially better subspace. Additionally, in order to trade off the exploration and exploitation, the framework involves a global search stage developed by combining leave-one-out cross-validation and radial basis function surrogate model. A performance selector is designed to switch the search dynamically and automatically between the global and local search stages. The empirical results on a variety of benchmark problems demonstrate that the proposed framework significantly outperforms several state-of-art algorithms with extremely limited fitness evaluations. Besides, the efficacy of Voronoi-diagram is furtherly analyzed, and the results show its potential to optimize very expensive problems.

\end{abstract}


\section{Introduction}

The evaluation of solutions to some real-world problems could be difficult or very expensive, in terms of computational cost or money.
For instance, evaluating once a solution to the Navier-Stokes equations involves several hours of computational fluid dynamic simulation~\cite{ong2003evolutionary}.
% Therefore, it is often the case that only a limited actual fitness evaluations (FEs) is available for searching for a solution. How to search efficiently toward an optimal solution within a small number of evaluations is a crucial problem.
In the past decades, evolutionary algorithms (EAs) have been widely applied to many real-world problems~\cite{jin2018data, guo2017robust}, but common EAs is not suitable to solve expensive problems due to the large number of FEs required to obtain an acceptable solution.
Hence, surrogate-assisted evolutionary algorithms (SAEAs), which sometimes sample from a cheap surrogate model instead of successively calling the actual complex evaluation process in conventional EAs \cite{jin2011surrogate}. 

SAEAs have been popular over the past years for its lower number of FEs required for convergence thanks to the use of surrogate models. Many effective models have been employed to reduce the number of FEs required. Examples include polynomial regression (PR)~\cite{zhou2005study}, Kriging model ~\cite{liu2014gaussian}, support vector machine (SVM)~\cite{xiang2017adaptive}, radial basis function (RBF)~\cite{yu2018surrogate} and neural network (NN)~\cite{sun2015two}. Besides, effective management strategies, like individual-based model management~\cite{jin2005comprehensive}, play a significantly important role in guaranteeing the convergence of SAEA~\cite{jin2011surrogate}.

Although SAEAs require fewer FEs to converge than common EAs and variations of SAEAs have shown their strength in solving some real-world CEPs (e.g., engineering design~\cite{koziel2014numerically}, health services~\cite{wang2016data}, interactive design~\cite{yan2013new}, etc.), the FEs required is still problematic in some very expensive problems.
Take the aircraft design as an example, the duration of accomplishing one crash simulation varies from $36$ to $160$ hours~\cite{wang2007review}.
If the SAEA starts to converge to a global optimal solution after $100$ FEs, the total optimization time will be between $150$ and $660$ days.
Despite the importance and demand of handling the budget issue, there is a lack of work considering very limited budget in the literature.
For example, Lu et al.~\cite{lu2014new} applied differential evolution (DE) assisted by rank-SVM for expensive problems given 100$D$ FEs as budget, where $D$ denotes the problem dimension. The Gaussian process assisted EA proposed by Liu et al.~\cite{liu2014gaussian} consumed fewer, but still, 50$D$ FEs. More recently, the particle swarm optimization (PSO) assisted by semi-supervised learning (SSLAPSO)~\cite{sun2018semi} and the active learning based PSO~\cite{wang2017committee} were both given only 11$D$ FEs as optimization budget.
 
All of the existed works use a budget of at least 11$D$ FEs probably due to the minimal number of FEs required to initialize the corresponding algorithm, otherwise these algorithms are not applicable any more.
In this work, we consider the extremely expensive problems with only severely limited actual fitness evaluations available and propose a new optimization algorithm named Voronoi-based efficient surrogate-assisted evolutionary algorithm. 
This work provides the following novel contributions:
\begin{itemize}
    \item To the best of our knowledge, this work is the first to solve very expensive black-box optimization problem with only 5$D$ FEs. With such a limited computational budget, the preference region and model accuracy of corresponding area are much important that we should devote the limited resource to potential optimal area.
    \item We proposed an efficient algorithm for very expensive optimization problem. Leave-one-out cross validation is employed to detect the uncertain area where the current surrogate model is not able to describe much precisely. Meanwhile, the Voronoi diagram is used to partition the search space and force the algorithm to exploit a relatively better area, which makes the algorithm more efficient for very expensive problems with limited fitness evaluations.
\end{itemize}

The remainder of paper is structured as follows. The literature review for SAEAs and the motivation for this work is introduced in Section \ref{realted-work}. Section \ref{methodology} describes the proposed framework and gives some further discussion on the new algorithm. Experiments are presented in \ref{experiment} for comparing the proposed algorithm with the state-of-art algorithms on a set of benchmark problems. Finally, a brief conclusion and the discussion of future work are drawn in Section \ref{conclusion}. 

\section{Literature Review}\label{realted-work}
Generally, the SAEAs for for CEPs \cite{lim2010generalizing} which could categorized into two kinds of frameworks in the literature, reducing the computational cost in different levels.

The first popular framework of SAEAs is inherited from the canonical evolutionary algorithm, which also contains mutation, crossover and selection operator \cite{jin2011surrogate}. However, the fitness evaluation consumption is still very huge because the involved population-based algorithm has to evaluate a certain number of individuals to guarantee the convergence and the diversity of the algorithm. For example, the classification- and regression-assisted differential evolution (CRADE) in \cite{lu2012classification} requires 10000 FEs for 30 dimension problems, which is not affordable in many real expensive problems.

Different from the framework discussed above, another SAEA framework is inherited from the efficient global optimization (EGO) \cite{jones1998efficient}. This kind of framework samples the solution in the whole search space, considering the uncertainty of surrogate model and the quality of the re-evaluated solution simultaneously, which aims to balance the exploitation and exploration in the optimization process.

Recently, researchers proposed different criteria to describe the model's uncertainty. For example, Wang et al.~\cite{wang2017committee} employed the query by committee active learning to assist this kind of SAEA (CALSAPSO), where the solution with the biggest disagreement among various models is regarded as the most uncertain solution.

% And the local search focuses on the surrogate model constructed by the top 10\% evaluated solutions to exploit the current best area.

Although the optimization ability of this category of SAEAs is significantly inferior to the first category of SAEAs, the fitness evaluation is dramatically reduced to obtained an acceptable solution. For example, the CALSAPSO only consumed $11D$ FEs to solve the expensive problem. Obviously, this kind of SAEAs is more appropriate for the expensive problem. 

In this paper, we focus on very expensive problems that the number of fitness evaluations is only setting as $5D$ in this paper. Obviously, the SAEA based on EGO framework is more appropriate for this problem. However, CALSAPSO, as the most efficient algorithm in the literature, is hard to handle very expensive problems. The ensemble model to determine the most uncertain solution could only help to improve the accuracy of the surrogate model. As a sequence, the number of fitness evaluation for exploitation will be very small that the performance for very expensive problem will be hindered. Furtherly, the CALSAPSO collects the top 10\% best evaluated solutions to build a surrogate model for local search, which only focuses on the region that the selected solutions determine. It has a big problem that the global optimum is likely to locate outside the local area and then the local search is not able to find a better solution for this case. 

To address the limitation of CALSAPSO and make algorithm more efficient for very expensive problems, we employ the leave-one-out cross-validation (LOOCV) and Voronoi diagram to embed in CALSAPSO's backbone. The LOOCV can detect the uncertain area of landscape and at the same time it might obtain a better solution. The Voronoi diagram, which has been demonstrated to be effective in robust optimization \cite{kevin2018voronoi}, provides a promising area for local search in the proposed algorithm. The detail of the proposed framework will be introduced in the next section.

\section{Voronoi based Efficient Surrogate-assisted Evolutionary Algorithm}\label{methodology}

\subsection{Overall framework}
\begin{figure*}[htbp]
    \centering
    \includegraphics[scale=0.65]{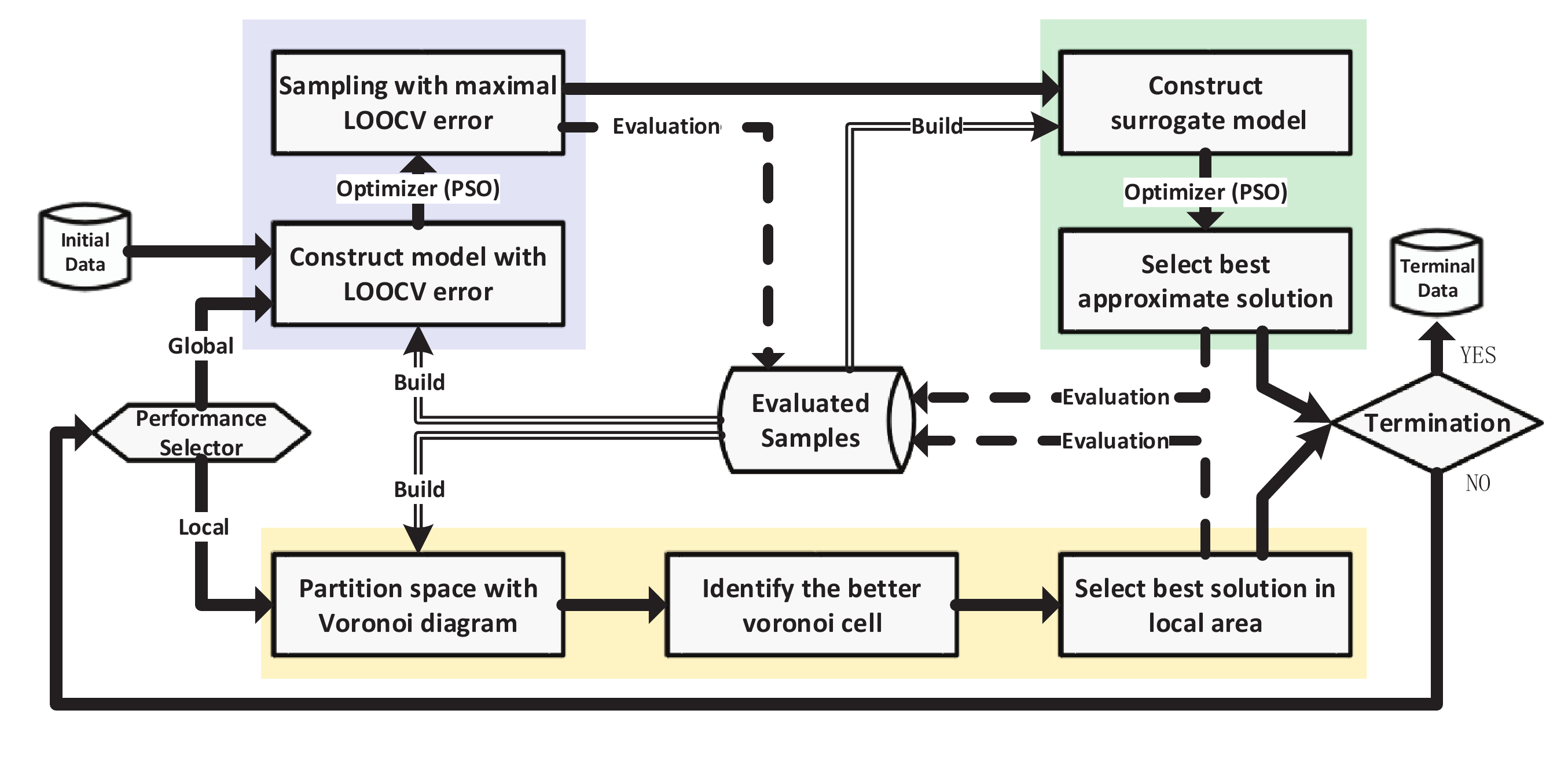}
    \caption{Generic framework of Voronoi-based efficient SAEA: The dashed lines represent to re-evaluate samples and then save into archive. Double solid lines denote to build surrogate model by evaluated samples in the archive.}
    \label{framework}
\end{figure*}

According to the previous discussion, we proposed a novel model management strategy to assist evolutionary algorithms, named as Voronoi-based efficient surrogate-assisted evolutionary algorithm (VESAEA). The overall procedure of VESAEA is presented in Fig. \ref{framework}. 

VESAEA initializes a dataset by Latin hypercube sampling (LHS) \cite{viana2016tutorial}, at first. Due to the limited fitness evaluations, we use part of fitness evaluations for initialization, which is set as $2D$ samples in this work. And then the algorithm operates the global search with evaluating two samples on the basis of LOOCV error and RBF surrogate model. The Voronoi-based local search will be applied if the global search makes no improvement during the optimization process according to a performance selector and vice versa. Finally, the algorithm will terminate if it runs out the total FEs setting as $5D$ in our experiments.

% We embed LOOCV error surrogate model and Voronoi diagram into the framework, which contains two search stages as the global search and the local search. The global search is made up of two stages: one employs the LOOCV error model to re-evaluate the most valuable individual and another aims to re-evaluate the most promising individual. On the other hand, the local search applies Voronoi diagram to select the best sample in the selected local area. Two search processes are controlled by a performance selector which switches the search stage to another one if the current search has no improvement in the previous generation.

In the framework, particle swarm optimization \cite{shi1998modified} is employed as the optimization algorithm in the global search process. All surrogates used in the framework are RBF model \cite{forrester2009recent}.

\subsection{Global search with LOOCV}

\paragraph{Construct model with LOOCV error} The LOOCV is able to measure the sensitivity of landscape and points with high LOOCV error usually located at the rugged area \cite{aute2013cross}. Therefore, re-evaluating samples with high LOOCV error will not only help to improve the model accuracy but also might discover the local optimal located in the rugged region. 

We could easily calculate the LOOCV error for evaluated samples as presented in Eq. \eqref{loocv}, where $\hat{y_i}$ is the approximate fitness value according to the surrogate model constructed excluding point $(\mathbf{x}, y_i)$.
\begin{equation}
    e(\mathbf{x_i}) = | y_i - \hat{y_i} |
    \label{loocv}
\end{equation}
However, it is impossible to obtain the LOOCV error for unobserved samples in search space. Suggested by \cite{aute2013cross}, we construct a surrogate model by LOOCV error of evaluated samples to approximate the LOOCV error of every point in the design landscape. Assume the dataset: 
\begin{equation*}
    \mathcal{D} = \{ (\mathbf{x}_1, y_1), (\mathbf x_2, y_2),..., (\mathbf x_N, y_N) \}
\end{equation*}
And by leave-one-out experiments, we could obtain the dataset of samples with LOOCV errors as:
\begin{equation*}
   \mathcal{E} = \{ (\mathbf{x}_1, e_{\mathbf{x}_1}), (\mathbf x_2, e_{\mathbf x_2}),..., (\mathbf x_N, e_{\mathbf x_N}) \} 
\end{equation*}

\paragraph{Sampling with maximal LOOCV error}
After completing all leave-one-out cross validation experiments, the surrogate model $\hat e_{LOO}(\bf x)$ will be built for LOOCV error. The RBF surrogate model with thin plate spline \cite{forrester2009recent} is used both in the cross-validation experiments and LOOCV error model construction. Once the model is constructed, the PSO optimizer will be applied to search for the point with the highest approximate LOOCV error in the whole landscape as presented in the Eq. \eqref{stage1}
\begin{equation}\large
    \mathbf{x}^* = \argmax_{\mathbf{x} \in S} {\hat e_{LOO}(\mathbf{x})}
    \label{stage1}
\end{equation}
When PSO satisfies the terminal condition, the solution with the highest LOOCV error found by PSO will be re-evaluated by the actual fitness function and be saved into an archive. 

\paragraph{Construct surrogate model}The second stage in global search is to optimize the global surrogate model. The global surrogate model $\hat f(\mathbf{x})$ is still built by the RBF model and optimized by the PSO algorithm. Therefore, the objective function changes into Eq. \eqref{stage2}: 
\begin{equation}\large
    \mathbf{x} = \argmax_{\mathbf{x} \in S} {\hat f(\mathbf{x})}
    \label{stage2}
\end{equation}
\paragraph{Select best approximate solution}
Once the PSO optimization process completing, the best solution found is re-evaluated using actual expensive fitness function and then the new sample will be added to the archive. 

\subsection{Voronoi based local search}
\begin{algorithm} \small
    \caption{Pseudo Code of Local Search}\label{local_search}
    {
        \KwIn{Samples: $\mathcal{S} = \{\mb{x}_i | i =1,2, ..., N\}$}
        % Partition search space by Voronoi diagram; 
        % Sort $\mathfrak{D}_{dataset}$ 
        $ \mathcal{P}_{rand} \leftarrow $ Generate a large number of samples in the search space; \\
        $n \leftarrow$ Dimension of the problem; \\
        $|\mathcal{P}_{rand}| = |\mathcal{S}| \times n \times 1000$ \\
        Each existed point $\mb{x}_i$ constructs one cell $C_i$; \\
        \For {$p_r \in P_{rand}$}
        {
            $p_r$ is assigned to the closest Voronoi cell $C_i$.
        }
        Voronoi cells $V = \{C_1, C_2, ..., C_N\}$ \\
        Identify the top 10\% best cells as $C_{top}=C_1 \cup C_2 ... \cup C_k$ \\
        Sample $\mathbf{x}^* = \argmin_{\mathbf{x}\in C_{top}} \hat f(\mathbf{x})$ \\
        \KwOut{New sample: $\mathbf{x}^*$}
    }
\end{algorithm}
In mathematic, the Voronoi diagram is used to partition the space into several small regions~\cite{garud2017design}. In this framework, we regard the evaluated sample of each Voronoi cell as the representative point and the Voronoi cells with better representative points are regarded as better cells. Then, the local search is operated in better cells. The procedure of Voronoi for local search is presented in Algorithm \ref{local_search}. 

\paragraph{Partition space with Voronoi diagram}
The landscape is firstly partitioned by samples in the sample set. Consider a sample set $\mathcal{S} = \{\mb{x_1}, \mb{x_2},\dots, \mb{x_N}\}$, each point constructs a Voronoi cell $V_i$ defined in \eqref{Voronoi}:
\begin{equation}
    %V_i = \{p_i \in \mathcal{S} | d(p, p_i) \leq d(p, p_j), \forall j \neq i\}
    V_i = \{\mb{x} \in \mathcal{S} | d(\mb{x}, \mb{x_i}) \leq d(\mb{x}, \mb{x_j}), \forall j \neq i\},
    \label{Voronoi}
\end{equation}
where, $d(\mb{x}, \mb{x'})$ denotes the Euclidean distance between points $\mb{x}$ and $\mb{x'}$~\cite{garud2017design}. Points of each Voronoi cell are closest to the corresponding evaluated sample of this cell. Thus, the boundary between the cells of any pair of adjacent points is the perpendicular bisector of these two points. Due to the irregular shape of Voronoi cell, it is hard to describe the boundaries with one specific equation. The Monte Carlo (MC) simulation is an alternative method to approximately identify the boundaries~\cite{xu2014robust} which is presented in Lines 1-8 of Algorithm \ref{local_search} where $|\cdot|$ denotes the size of one sample set.

\paragraph{Identify the better Voronoi cell}
The top 10\% best cells are selected to form a pool of samples $C_{top}$ as presented in Line 9 in Algorithm \ref{local_search}. Assume the number of 10\% best cells and total random samples are $k$, $NP$ respectively. Then, the $C_{top}$ can be described as Eq. \eqref{ctop}. 
\begin{equation}
    C_{top} = C_1 \cup C_2 ... \cup C_k = \{p_1, p_2, ..., p_{NP}\}
    \label{ctop}
\end{equation}

\paragraph{Select best solution in local area}
Finally, the framework will select the best sample in the set of $C_{top}$ according to the approximate model $\hat f(\mathbf{x})$ that the objective function is described as Eq. \eqref{local_sample}. The new sample will also be re-evaluated with expensive fitness function after finishing local search and then added to the archive. 
\begin{equation}
    \mathbf{x}^* = \argmin_{x\in C_{top}} \hat f(\mathbf{x})
    \label{local_sample}
\end{equation}

\begin{figure}
    \centering
    \includegraphics[scale=0.5]{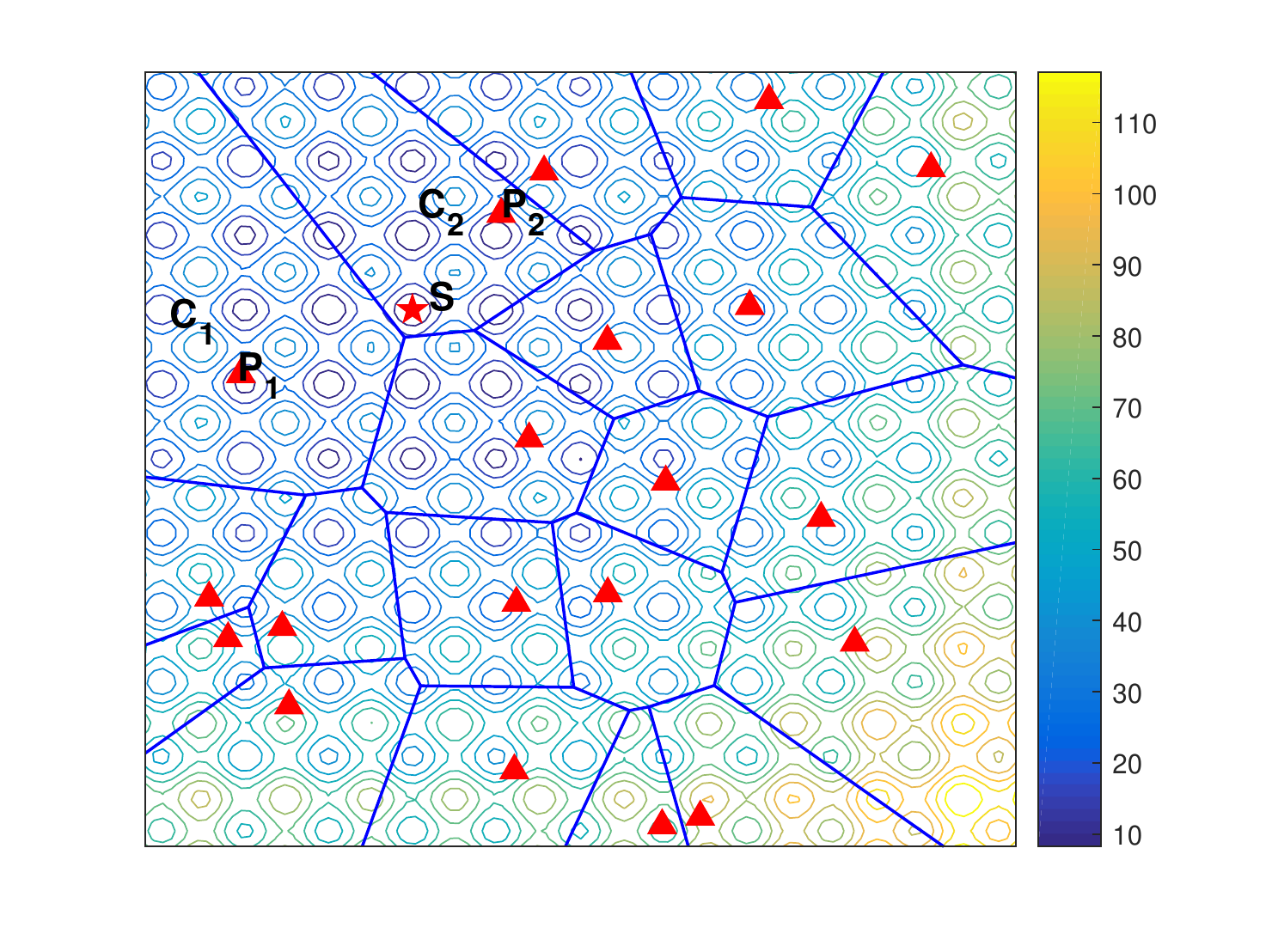}
    \caption{An example of Voronoi for local search: triangles are evaluated samples, the star is the optimal solution, lines are boundaries of Voronoi cell.}
    \label{VLS}
\end{figure}

Take Fig. \ref{VLS} as an example, $P_1, P_2$ are two best samples so that $C_1, C_2$ are two best Voronoi cells and $S$ is the optimal solution. According to the local search strategy, the new sample will be selected from $C_{top} = C_1 \cup C_2$. As shown in Fig. \ref{VLS}, we could easily find the optimal solution is likely to locate in few top best Voronoi cells. 

\subsection{Discussion}
\paragraph{Search behavior}
In VESAEA, a performance selector is applied to trade off the exploration and exploitation. The global search explores potentially better area in the whole search space and then the local search is applied to exploit the specific area to obtain a local optimum if global search has no improvement in current generation.

The global search firstly starts with sampling with respect to the LOOCV error to increase the model's accuracy. After re-evaluating the sample with maximal approximate LOOCV error, the new surrogate model built by all samples in the dataset is more likely to describe the valuable rugged area of landscape where the local optimal might hidden. Then, re-evaluating the best individual of new accurate surrogate model might find a better solution.

By contrast, the local search considers exploiting the potential area. The space partition makes local search focus more on the specific better area and meanwhile, the top 10\% cells are selected to avoid being trap into a local optimum. Monte Carlo simulation is used to describe the Voronoi cells and it only requires to select the new solution among the generated random samples belonging to the selected cells, which avoids searching next sample by complex algorithms.

\paragraph{Computational complexity}
For very expensive problems, the cost of fitness evaluation is more expensive than algorithm's self. Besides, the most time-consuming part in the framework is Voronoi-based local search. In our framework, the number of random points generated for Monte Carlo simulation will be very large when the dimension of problem increases and the distance between generated points with every evaluated sample need to be calculated, whose computational complexity will be very high. Another time-consuming part is the construction of LOOCV error model, which need to construct surrogate model and calculate LOOCV error for every evaluated point. As a result, the framework is not suitable for large-scale expensive problem, in which the algorithm' complexity will be terribly huge.

\begin{table*}[htpb!]
\caption{The results of five algorithms: VESAEA, EGO-LCB, GPEME, SSL-APSO, CAL-SAPSO performed on 25 test functions over 25 runs including the average best fitness and standard deviation shown as AVR $\pm$ STD. The boldface figures are the best fitness among five algorithms in each test problem.} \label{result}        
\resizebox{\textwidth}{!}{        
\begin{tabular}{|c|c|c|c|c|c|c|}
\hline
Problem      & D    & VESAEA                        & EGO-LCB       & GPEME       & SSLAPSO    & CALSAPSO      \\ \hline
Sphere       & 5    & 8.49e+02 $\pm$ 6.23e+02        & \textbf{3.78e+02 $\pm$ 2.24e+02} & 2.05e+03 $\pm$ 1.62e+03 & 6.34e+03 $\pm$ 2.96e+03 & 1.28e+03 $\pm$ 1.15e+03     \\ \hline
Sphere       & 10   & \textbf{1.88e+03 $\pm$ 6.92e+02} & 5.73e+03 $\pm$ 2.79e+03     & 7.02e+03 $\pm$ 2.58e+03 & 1.71e+04 $\pm$ 5.67e+03 & 6.21e+03 $\pm$ 1.75e+03     \\ \hline
Sphere       & 15   & \textbf{4.85e+03 $\pm$ 7.64e+02} & 2.07e+04 $\pm$ 3.79e+03     & 1.65e+04 $\pm$ 4.86e+03 & 2.56e+04 $\pm$ 4.15e+03 & 1.13e+04 $\pm$ 3.01e+03     \\ \hline
Sphere       & 20   & \textbf{1.09e+04 $\pm$ 1.49e+03} & 4.41e+04 $\pm$ 7.05e+03     & 2.99e+04 $\pm$ 7.34e+03 & 4.42e+04 $\pm$ 7.62e+03 & 2.07e+04 $\pm$ 5.18e+03     \\ \hline
Sphere       & 30   & \textbf{2.57e+04 $\pm$ 2.55e+03} & 8.12e+04 $\pm$ 8.81e+03     & 5.04e+04 $\pm$ 1.15e+04 & 6.02e+04 $\pm$ 6.90e+03 & 2.99e+04 $\pm$ 8.19e+03     \\ \hline
Rosenbrock   & 5    & \textbf{5.46e+01 $\pm$ 3.00e+01} & 7.44e+01 $\pm$ 3.21e+01     & 1.64e+02 $\pm$ 1.22e+02 & 2.86e+02 $\pm$ 2.18e+02 & 1.28e+02 $\pm$ 1.21e+02     \\ \hline
Rosenbrock   & 10   & \textbf{2.38e+02 $\pm$ 1.02e+02} & 1.13e+03 $\pm$ 4.00e+02     & 1.08e+03 $\pm$ 6.19e+02 & 2.31e+03 $\pm$ 1.16e+03 & 3.76e+02 $\pm$ 2.25e+02     \\ \hline
Rosenbrock   & 15   & \textbf{4.69e+02 $\pm$ 1.74e+02} & 2.84e+03 $\pm$ 1.07e+03     & 1.47e+03 $\pm$ 8.97e+02 & 2.87e+03 $\pm$ 1.20e+03 & 5.57e+02 $\pm$ 5.25e+02     \\ \hline
Rosenbrock   & 20   & 9.36e+02 $\pm$ 2.41e+02   & 5.28e+03 $\pm$ 1.46e+03     & 3.17e+03 $\pm$ 1.36e+03 & 4.44e+03 $\pm$ 1.48e+03 & \textbf{5.00e+02 $\pm$ 1.88e+02} \\ \hline
Rosenbrock   & 30   & 2.42e+03 $\pm$ 4.35e+02   & 1.48e+04 $\pm$ 2.25e+03     & 7.58e+03 $\pm$ 2.79e+03 & 8.43e+03 $\pm$ 2.41e+03 & \textbf{9.47e+02 $\pm$ 6.45e+02} \\ \hline
Ackley       & 5    & 1.60e+01 $\pm$ 3.58e+00   & \textbf{9.11e+00 $\pm$ 2.22e+00} & 1.69e+01 $\pm$ 2.68e+00 & 1.93e+01 $\pm$ 1.36e+00 & 1.80e+01 $\pm$ 2.36e+00     \\ \hline
Ackley       & 10   & 1.66e+01 $\pm$ 1.89e+00   & \textbf{1.57e+01 $\pm$ 5.35e+00} & 1.91e+01 $\pm$ 1.47e+00 & 2.04e+01 $\pm$ 5.21e-01 & 1.98e+01 $\pm$ 5.49e-01     \\ \hline
Ackley       & 15   & \textbf{1.75e+01 $\pm$ 1.19e+00} & 1.99e+01 $\pm$ 9.31e-01     & 1.93e+01 $\pm$ 1.63e+00 & 2.05e+01 $\pm$ 3.58e-01 & 1.98e+01 $\pm$ 3.13e-01     \\ \hline
Ackley       & 20   & \textbf{1.84e+01 $\pm$ 7.36e-01} & 2.05e+01 $\pm$ 2.27e-01     & 1.98e+01 $\pm$ 1.10e+00 & 2.06e+01 $\pm$ 2.12e-01 & 1.99e+01 $\pm$ 3.81e-01     \\ \hline
Ackley       & 30   & \textbf{1.91e+01 $\pm$ 3.68e-01} & 2.08e+01 $\pm$ 1.76e-01     & 2.01e+01 $\pm$ 3.70e-01 & 2.05e+01 $\pm$ 1.24e-01 & 1.98e+01 $\pm$ 4.78e-01     \\ \hline
Griewank     & 5    & 9.85e+00 $\pm$ 6.09e+00   & \textbf{4.74e+00 $\pm$ 1.75e+00} & 1.97e+01 $\pm$ 1.28e+01 & 6.08e+01 $\pm$ 2.43e+01 & 2.05e+01 $\pm$ 1.02e+01     \\ \hline
Griewank     & 10   & \textbf{1.78e+01 $\pm$ 5.11e+00} & 4.69e+01 $\pm$ 2.04e+01     & 8.37e+01 $\pm$ 3.13e+01 & 1.45e+02 $\pm$ 4.51e+01 & 6.64e+01 $\pm$ 2.83e+01     \\ \hline
Griewank     & 15   & \textbf{5.04e+01 $\pm$ 1.04e+01} & 1.72e+02 $\pm$ 3.96e+01     & 1.53e+02 $\pm$ 4.65e+01 & 2.13e+02 $\pm$ 4.04e+01 & 1.11e+02 $\pm$ 2.50e+01     \\ \hline
Griewank     & 20   & \textbf{1.01e+02 $\pm$ 1.16e+01} & 3.86e+02 $\pm$ 5.88e+01     & 2.68e+02 $\pm$ 6.93e+01 & 3.69e+02 $\pm$ 4.67e+01 & 1.89e+02 $\pm$ 3.43e+01     \\ \hline
Griewank     & 30   & \textbf{2.27e+02 $\pm$ 1.55e+01} & 7.14e+02 $\pm$ 9.22e+01     & 4.67e+02 $\pm$ 1.12e+02 & 5.04e+02 $\pm$ 4.69e+01 & 2.92e+02 $\pm$ 6.80e+01     \\ \hline
Rastrigin    & 5    & 3.90e+01 $\pm$ 9.99e+00   & \textbf{3.51e+01 $\pm$ 8.14e+00} & 3.78e+01 $\pm$ 1.17e+01 & 5.57e+01 $\pm$ 1.35e+01 & 4.12e+01 $\pm$ 1.21e+01     \\ \hline
Rastrigin    & 10   & \textbf{8.15e+01 $\pm$ 1.49e+01} & 1.01e+02 $\pm$ 2.04e+01     & 8.71e+01 $\pm$ 1.78e+01 & 1.22e+02 $\pm$ 1.99e+01 & 8.57e+01 $\pm$ 1.68e+01     \\ \hline
Rastrigin    & 15   & 1.35e+02 $\pm$ 2.21e+01   & 1.90e+02 $\pm$ 2.03e+01     & 1.63e+02 $\pm$ 3.62e+01 & 2.03e+02 $\pm$ 2.21e+01 & \textbf{1.28e+02 $\pm$ 2.76e+01} \\ \hline
Rastrigin    & 20   & \textbf{1.66e+02 $\pm$ 3.48e+01} & 2.72e+02 $\pm$ 1.98e+01     & 2.35e+02 $\pm$ 2.96e+01 & 2.82e+02 $\pm$ 2.36e+01 & 1.79e+02 $\pm$ 4.36e+01     \\ \hline
Rastrigin    & 30   & \textbf{2.36e+02 $\pm$ 5.36e+01} & 4.67e+02 $\pm$ 2.90e+01     & 3.73e+02 $\pm$ 3.72e+01 & 4.33e+02 $\pm$ 2.50e+01 & 2.37e+02 $\pm$ 6.15e+01          \\ \hline
\multicolumn{2}{|c|}{Average ranking}  & 1.36                            & 3.44                            & 3.04                    & 4.72                  & 2.44                                                                        \\ \hline
\multicolumn{2}{|c|}{Adjusted p-value} & NA                              & 3.30E-05                       & 1.60E-03                 & 6.10E-13              & 0.1112                                                                      \\ \hline
% \multicolumn{2}{|c|}{Average ranking}  & 1.4167                            & 3.40                            & 3.00                    & 4.67                  & 2.47                                                                        \\ \hline
% \multicolumn{2}{|c|}{Adjusted p-value} & NA                              & 0.073                       & 0.0607                 & 3.00E-07              & 0.4141                                                                      \\ \hline

\end{tabular}
}
\end{table*}

\section{Numerical Experiment and analysis}\label{experiment}
In order to analyse the efficacy of VESAEA, we compared the proposed algorithm with several state-of-art algorithms on commonly used benchmark problems in this section. Then the empirical analysis of Voronoi diagram's effect are presented. The source code can be downloaded from the Github $\footnote{https://github.com/HawkTom/VESAEA}$. 

\subsection{Experimental Setup}
One efficient classical algorithm and three state-of-the-art algorithms are selected in our experiments, which have different characteristics on CEPs. A brief description of four algorithms is presented below.
\begin{itemize}
    \item \emph{EGO-LCB}: Lower confidence bound (LCB) criterion based efficient global optimization \cite{jones1998efficient} is a classical algorithm but performs efficiently on low-dimension expensive problems. 
    \item \emph{GPEME}\cite{liu2014gaussian}: DE assisted by Kriging model with individual-based evolution strategy. The LCB criterion is also employed as the re-evaluation strategy. 
    \item \emph{SSLAPSO}\cite{sun2018semi}: PSO is enhanced by semi-supervised learning which is employed to make use of both evaluated and un-evaluated samples. Two RBF surrogate models are built by true evaluated and approximately evaluated samples respectively to determine new individuals' fitness.  
    \item \emph{CALSAPSO}\cite{wang2017committee}: The committee-based active learning is applied to SAEA framework by using ensemble of PR, RBF and Kriging model. Three surrogate models are used simultaneously to determine the most uncertain samples and potentially best solutions. 
\end{itemize}

\begin{table}[htpb!]\tiny
    \centering
    \caption{Benchmark Problems. The optimum solution is shifted to another random position in the landscape.}\label{bench}
    \resizebox{\linewidth}{!}{
    \begin{tabular}{|c|c|c|c|}
    \hline
    Problem    & Dimension     & Optimum & Note                                                                       \\ \hline
    Sphere     & 5,10,15,20,30 & 0       & Uni-modal                                                                  \\ \hline
    Griewank   & 5,10,15,20,30 & 0       & Multi-modal                                                                \\ \hline
    Ackley     & 5,10,15,20,30 & 0       & Multi-modal                                                                \\ \hline
    Rosenbrock & 5,10,15,20,30 & 0       & Multi-modal with narrow valley                                             \\ \hline
    Rastrigin  & 5,10,15,20,30 & 0       & Very complicated multi-modal                                               \\ \hline
    \end{tabular}}                   
    \end{table}

The experiments are performed on 5 widely used benchmark problems with $D=5,10,15,20,30$ listed in Table \ref{bench}. The size of initial samples is set to $2D$ and all algorithms terminate after $5D$ real fitness evaluations. The surrogate models used in all algorithms are all imported from SURROGATES tool-box \cite{surrogatestoolbox2p1}, which is based on a RBF toolbox \cite{rbftoolbox}. 

\subsection{Comparative Experiment on Benchmark Problems}

\begin{figure*}[htpb!]
    \centering
    \includegraphics[width=.18\linewidth]{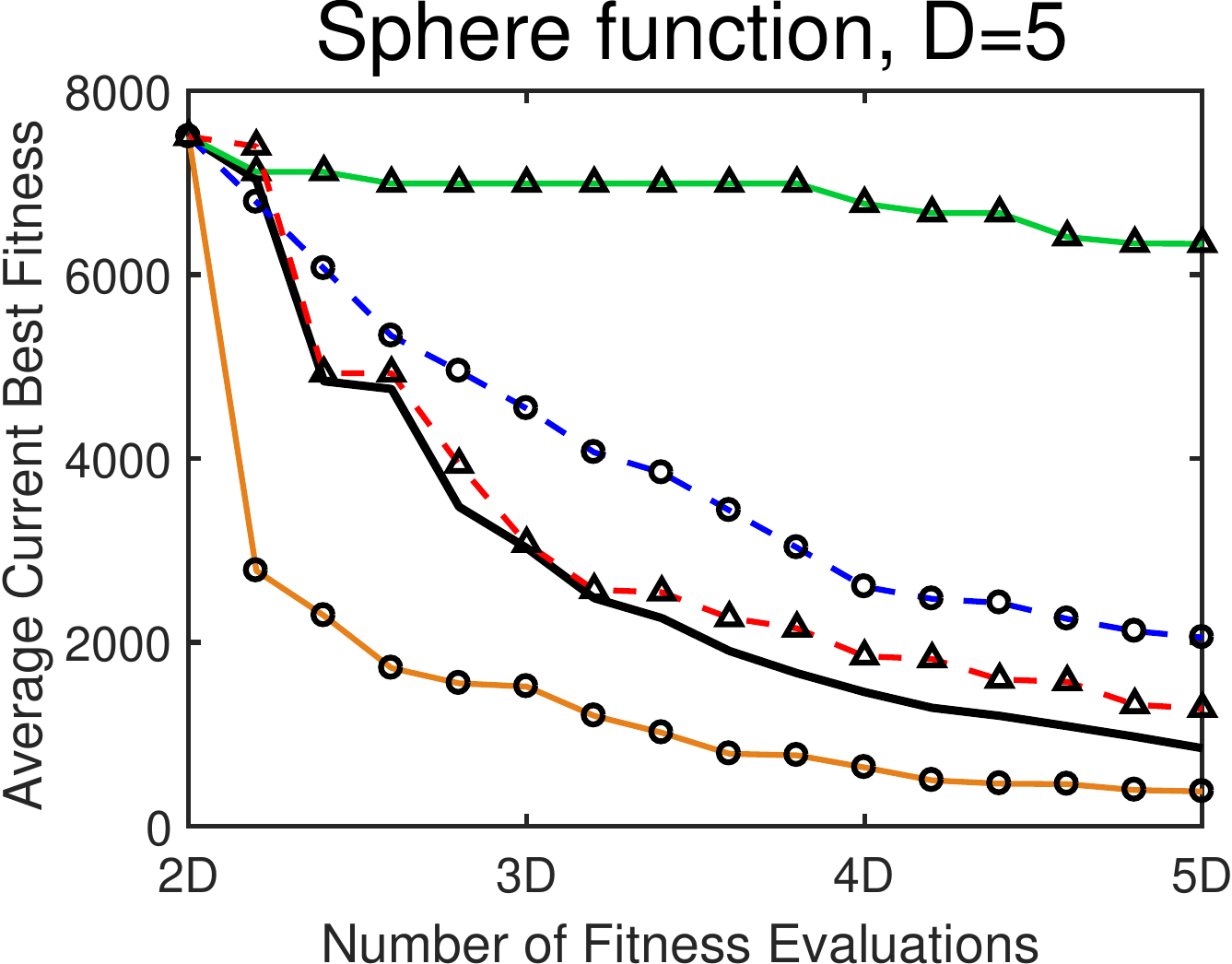}\quad
    \includegraphics[width=.18\linewidth]{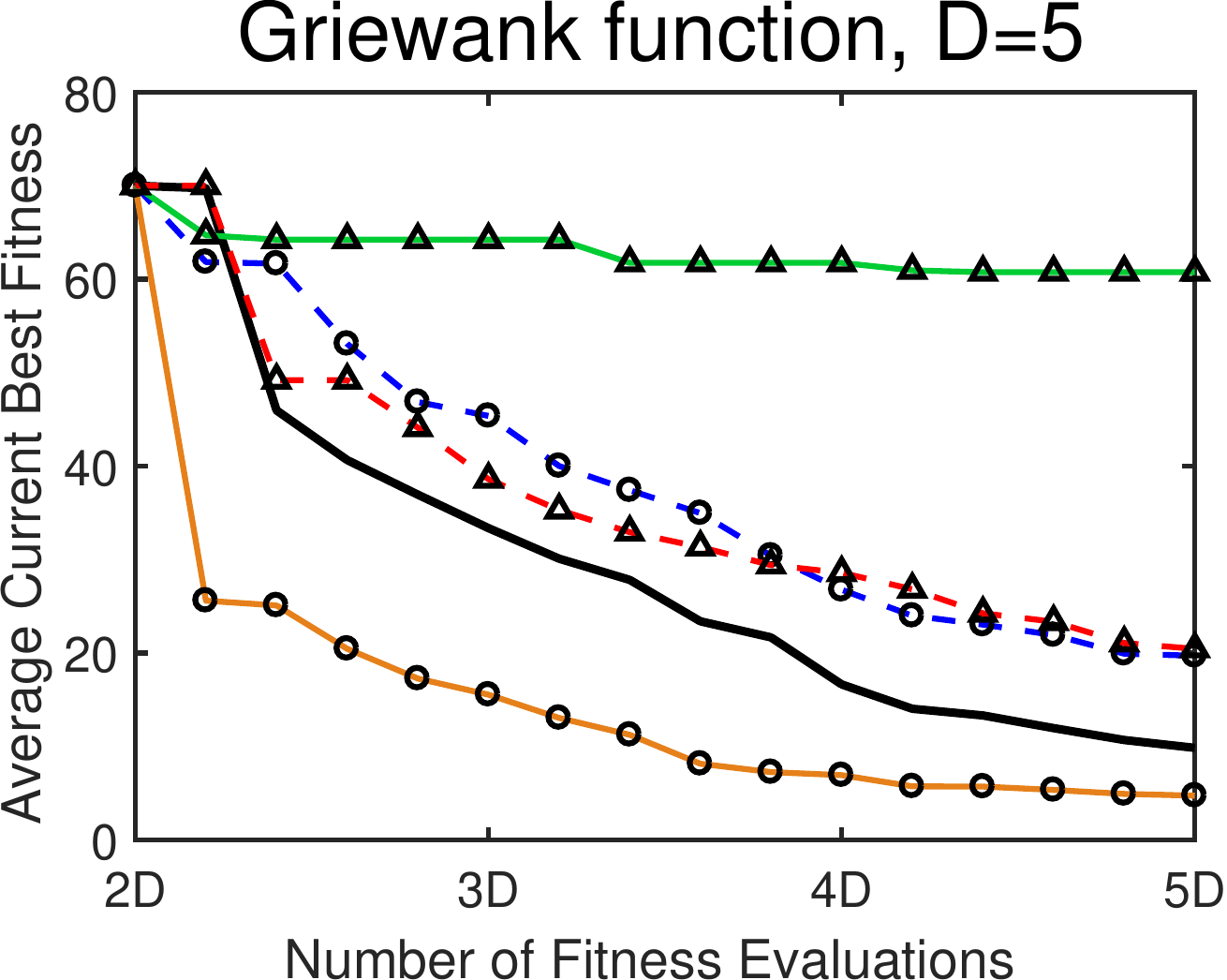}\quad    
    \includegraphics[width=.18\linewidth]{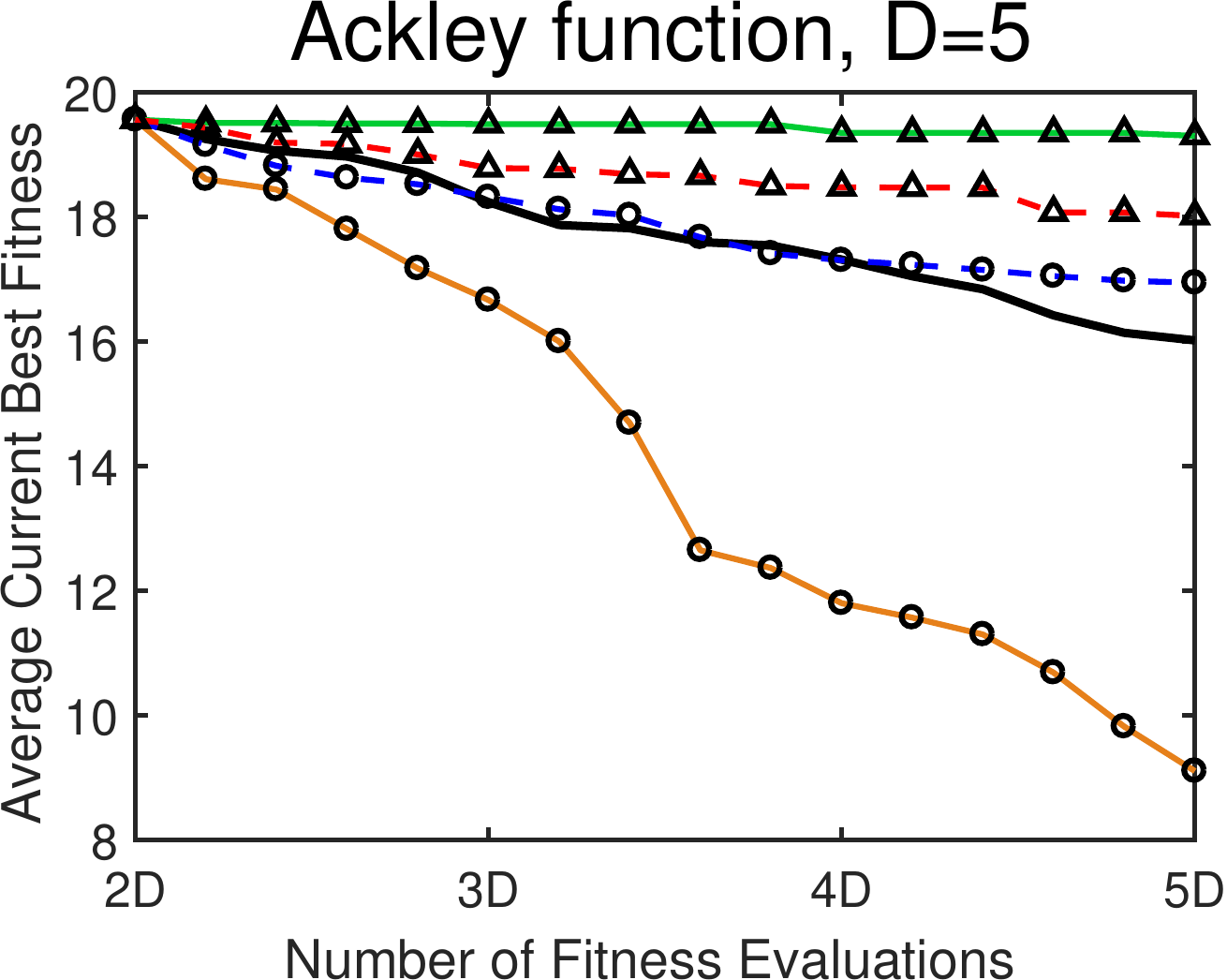}\quad
    \includegraphics[width=.18\linewidth]{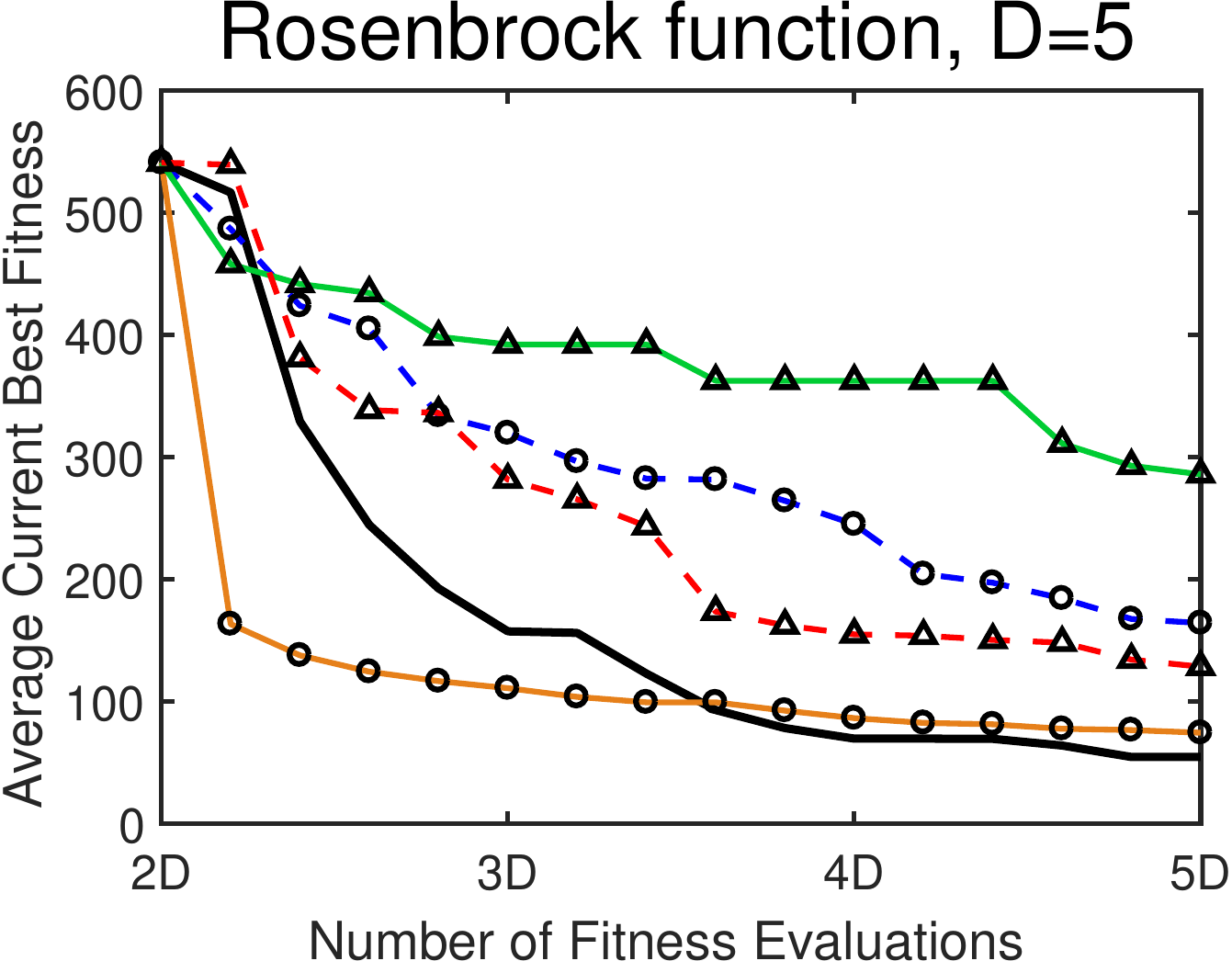}\quad 
    \includegraphics[width=.18\linewidth]{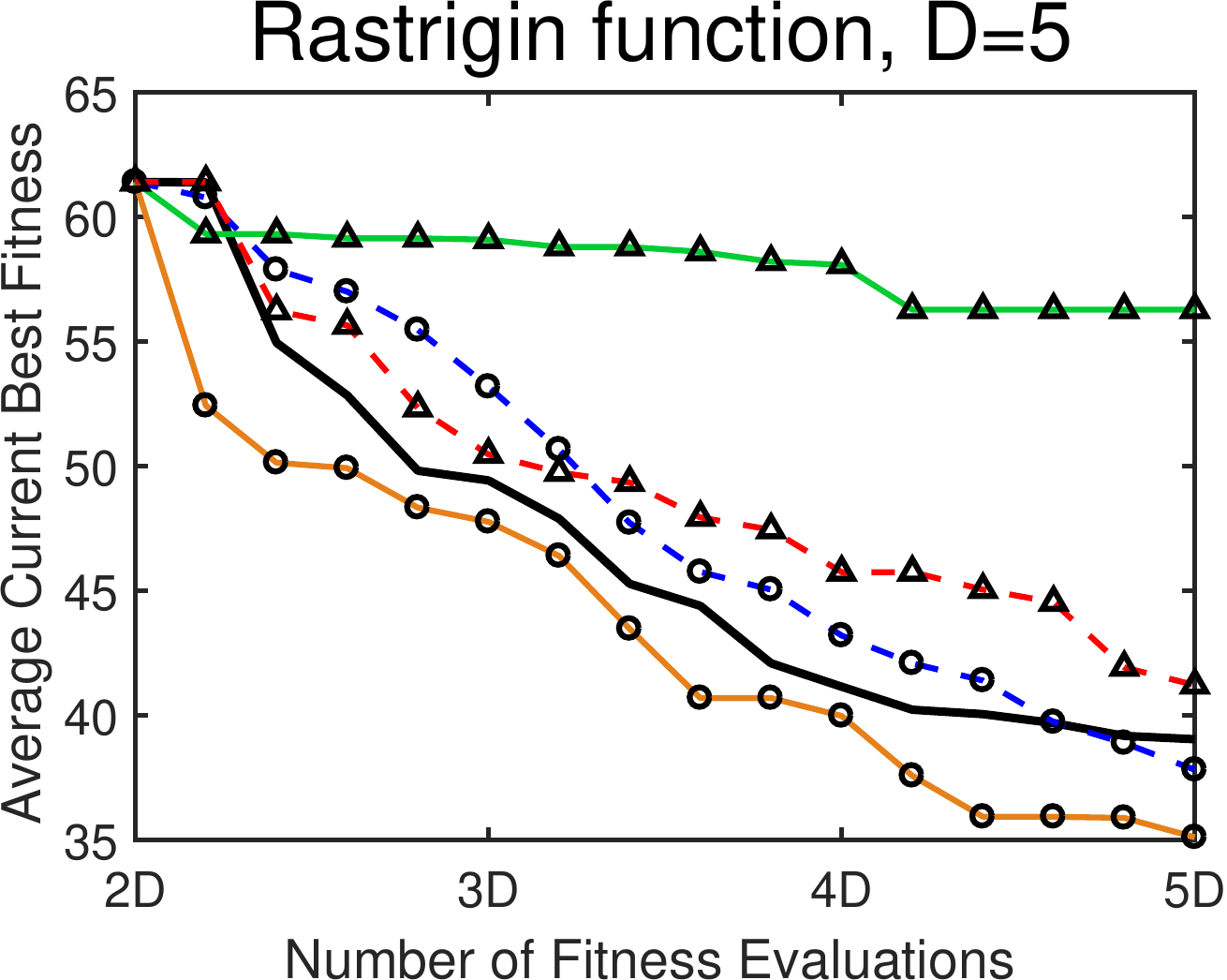}
    \medskip
    
    \includegraphics[width=.18\linewidth]{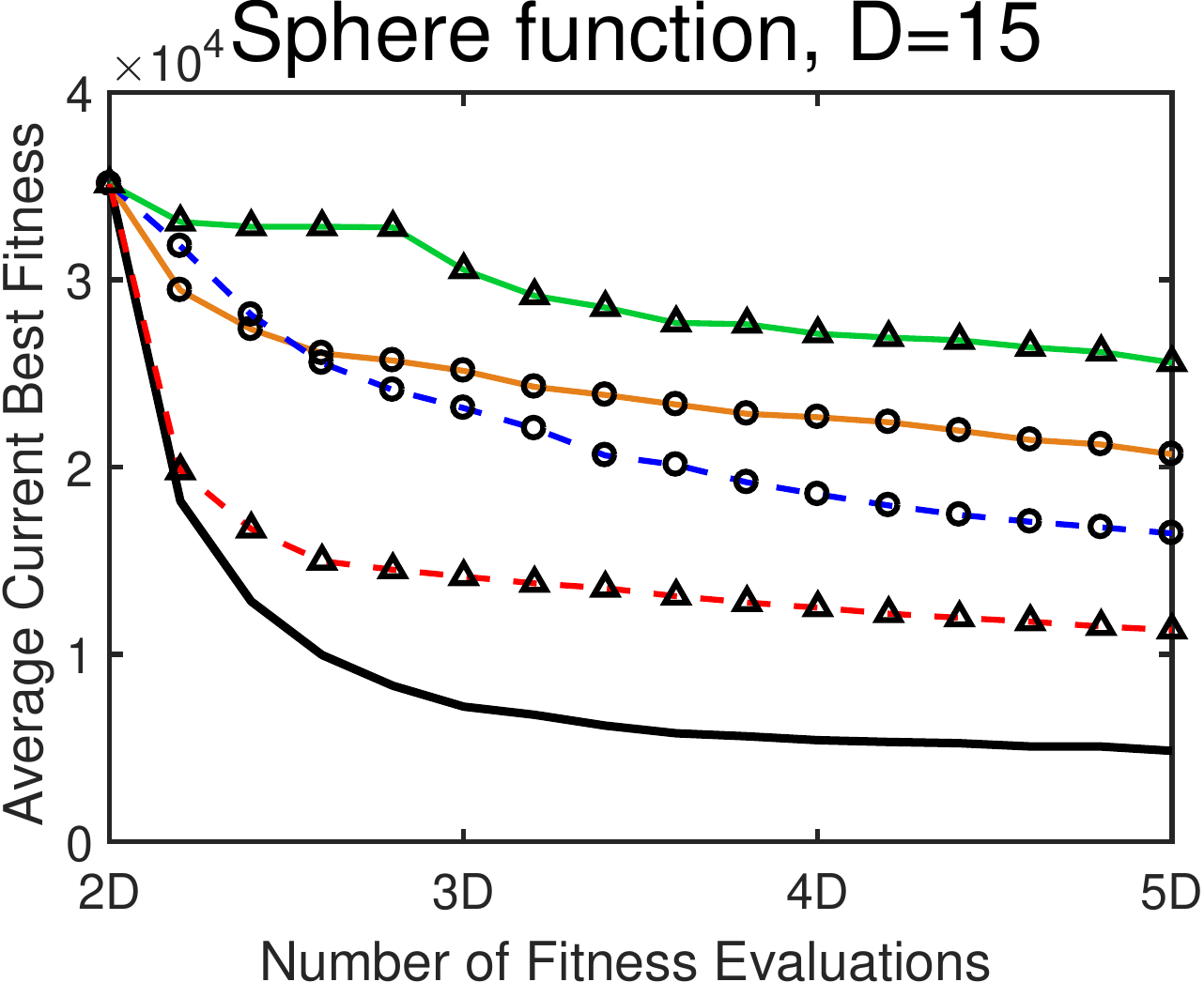}\quad
    \includegraphics[width=.18\linewidth]{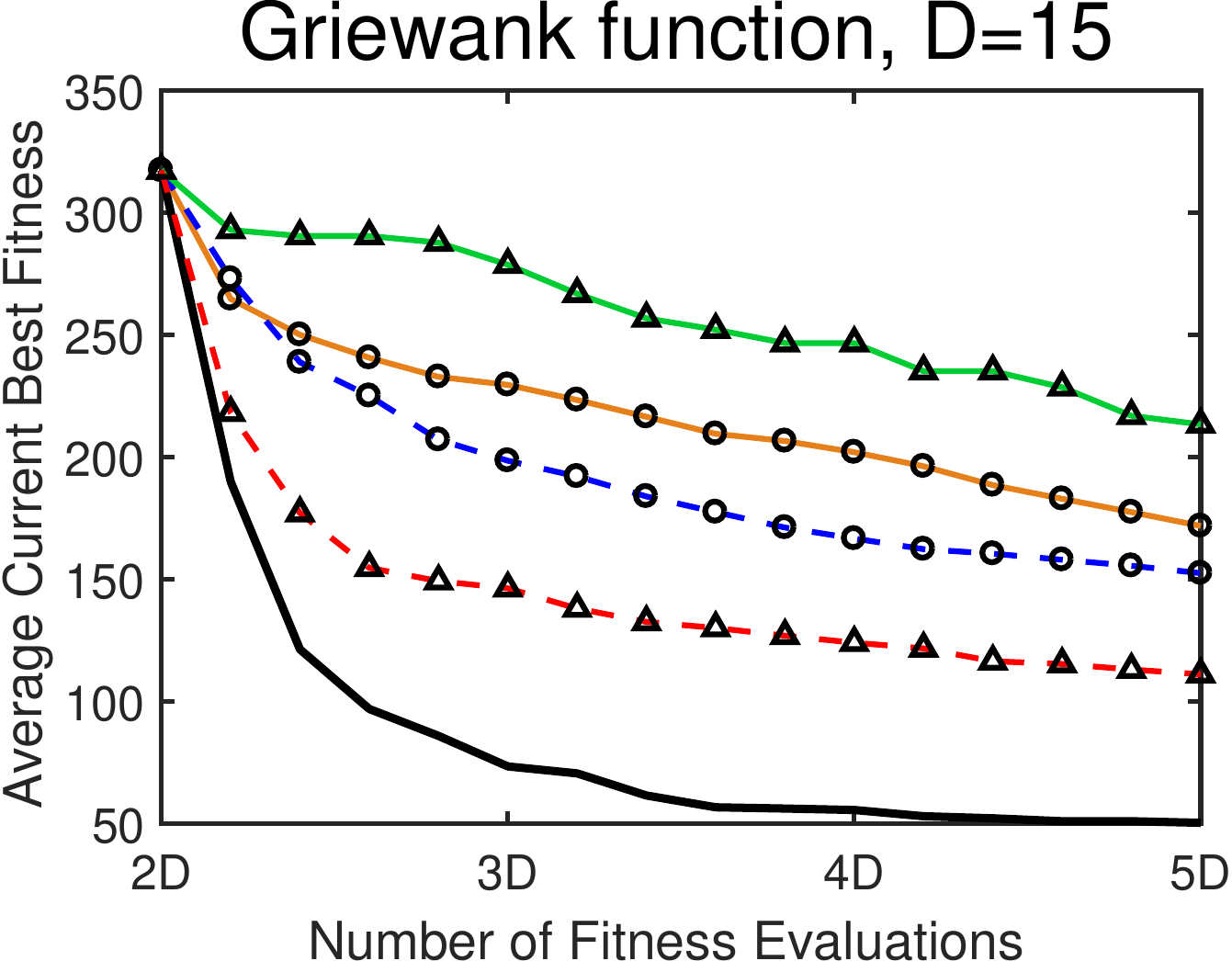}\quad    
    \includegraphics[width=.18\linewidth]{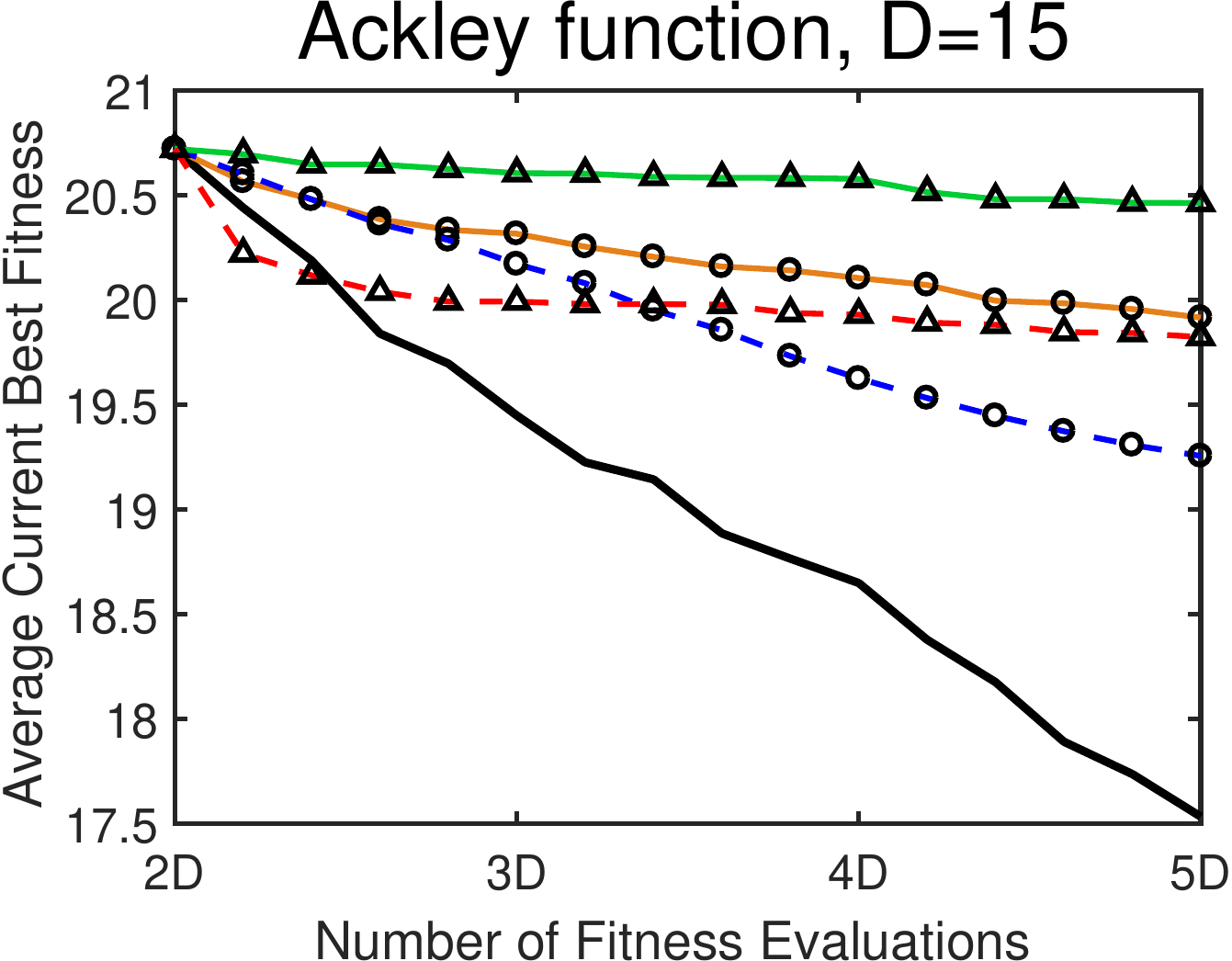}\quad
    \includegraphics[width=.18\linewidth]{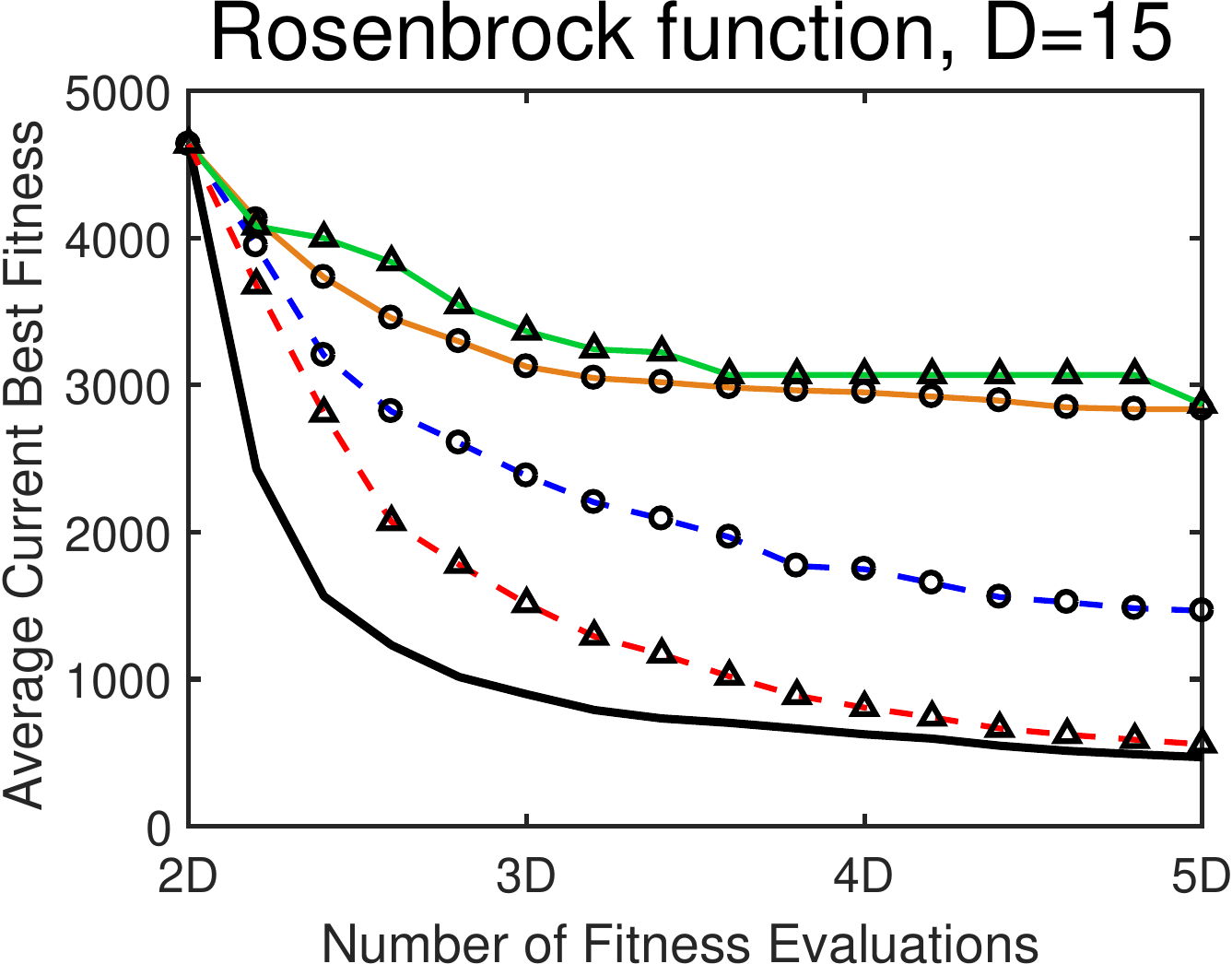}\quad 
    \includegraphics[width=.18\linewidth]{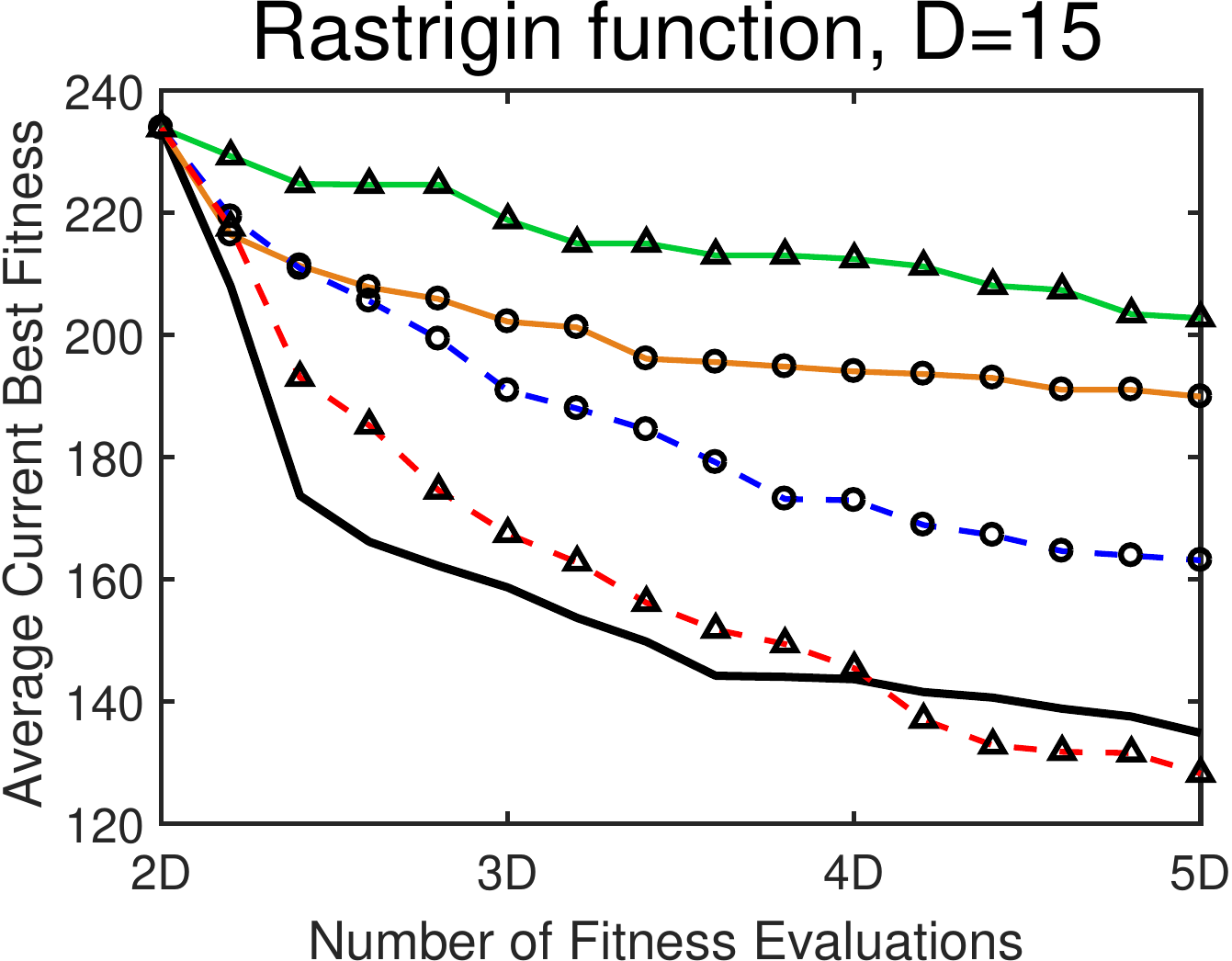}
    \medskip
    
    \includegraphics[width=.18\linewidth]{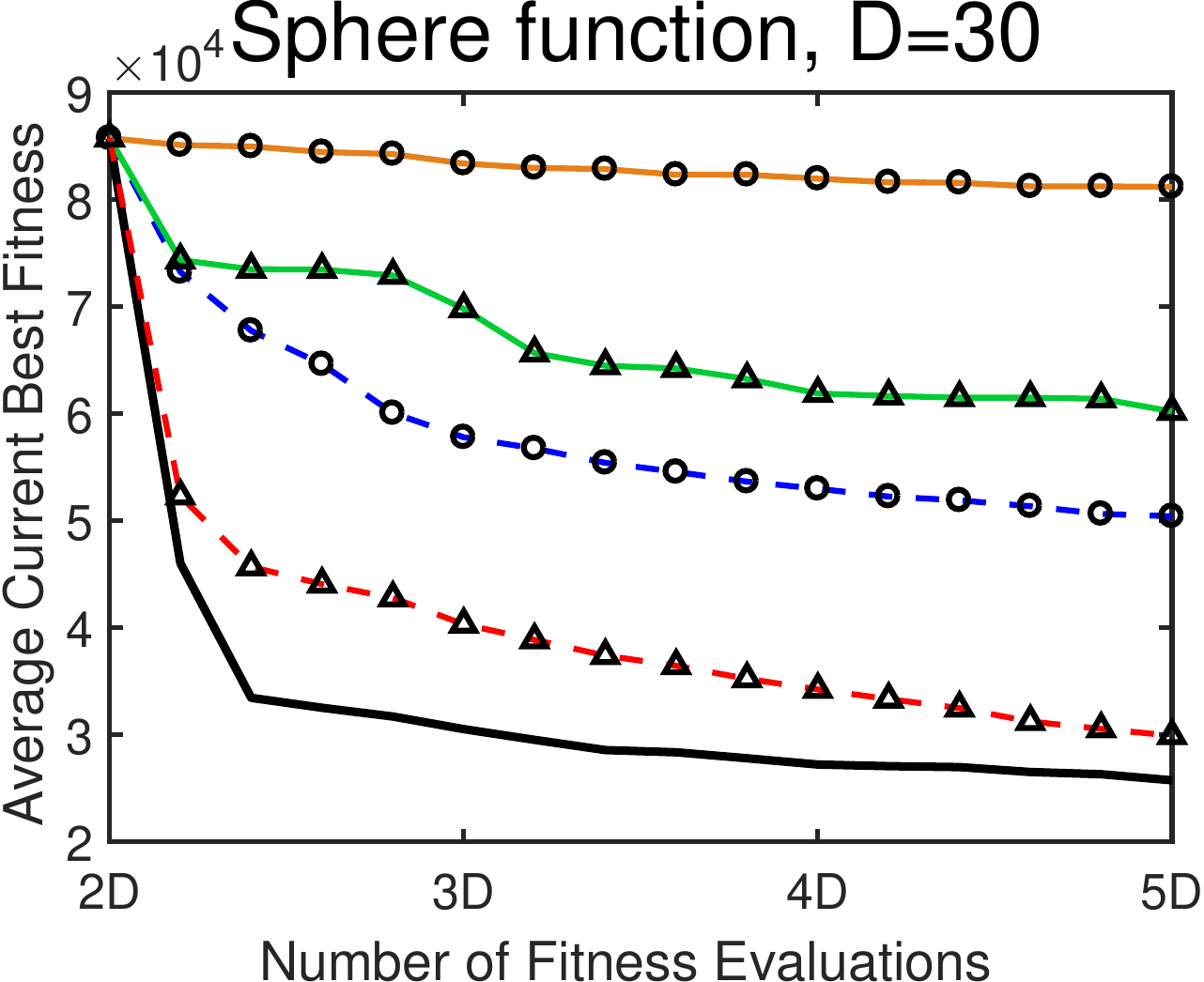}\quad
    \includegraphics[width=.18\linewidth]{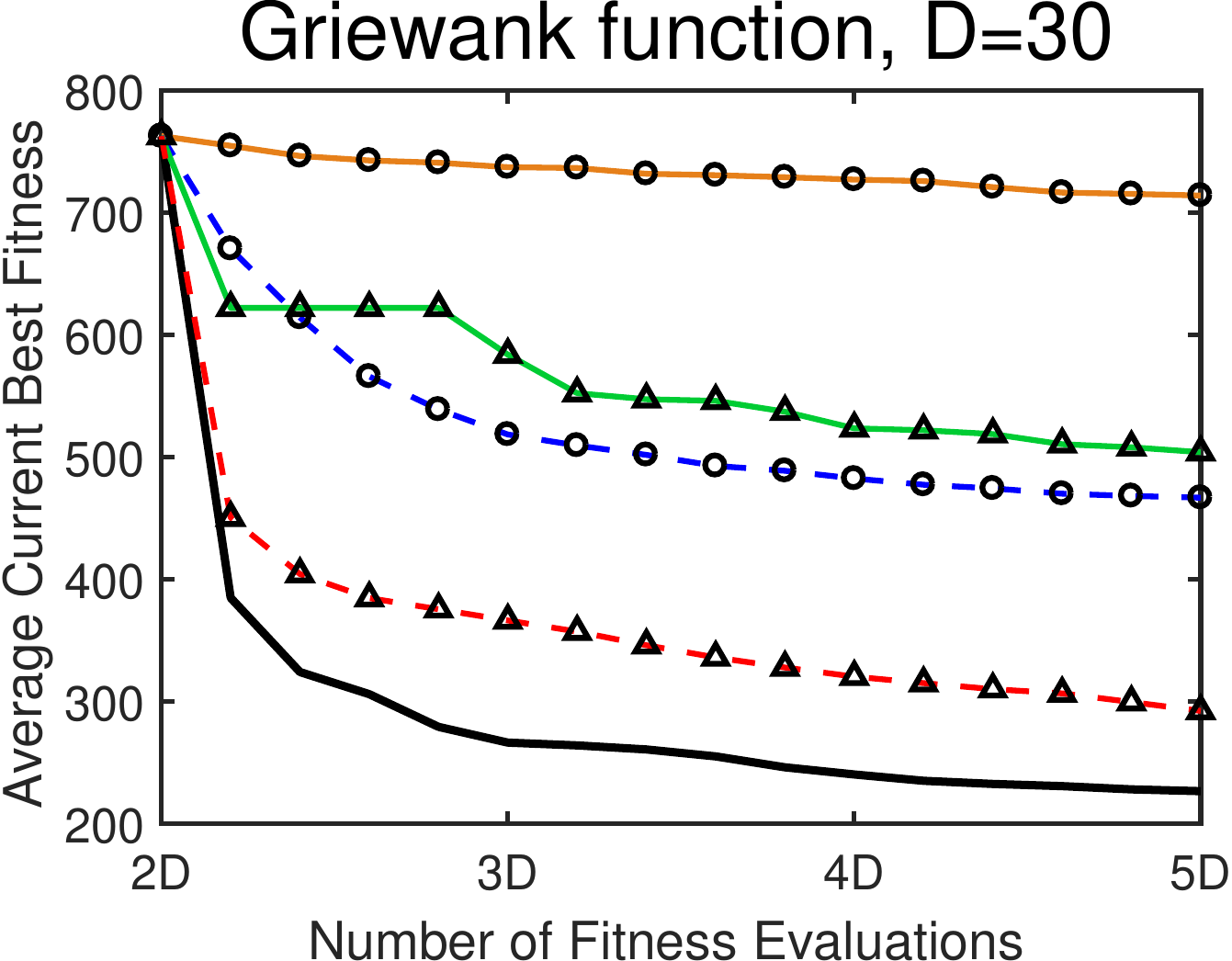}\quad    
    \includegraphics[width=.18\linewidth]{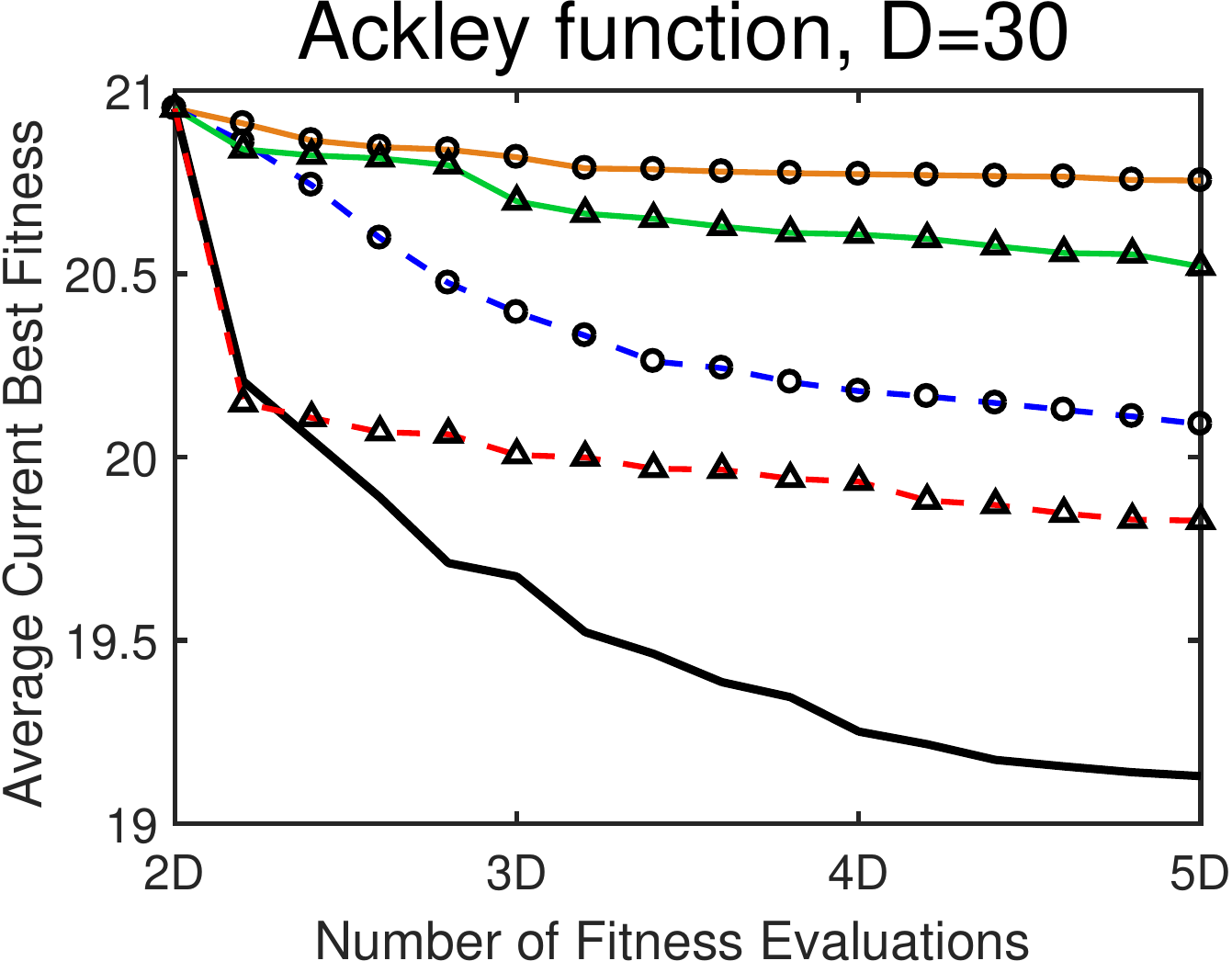}\quad
    \includegraphics[width=.18\linewidth]{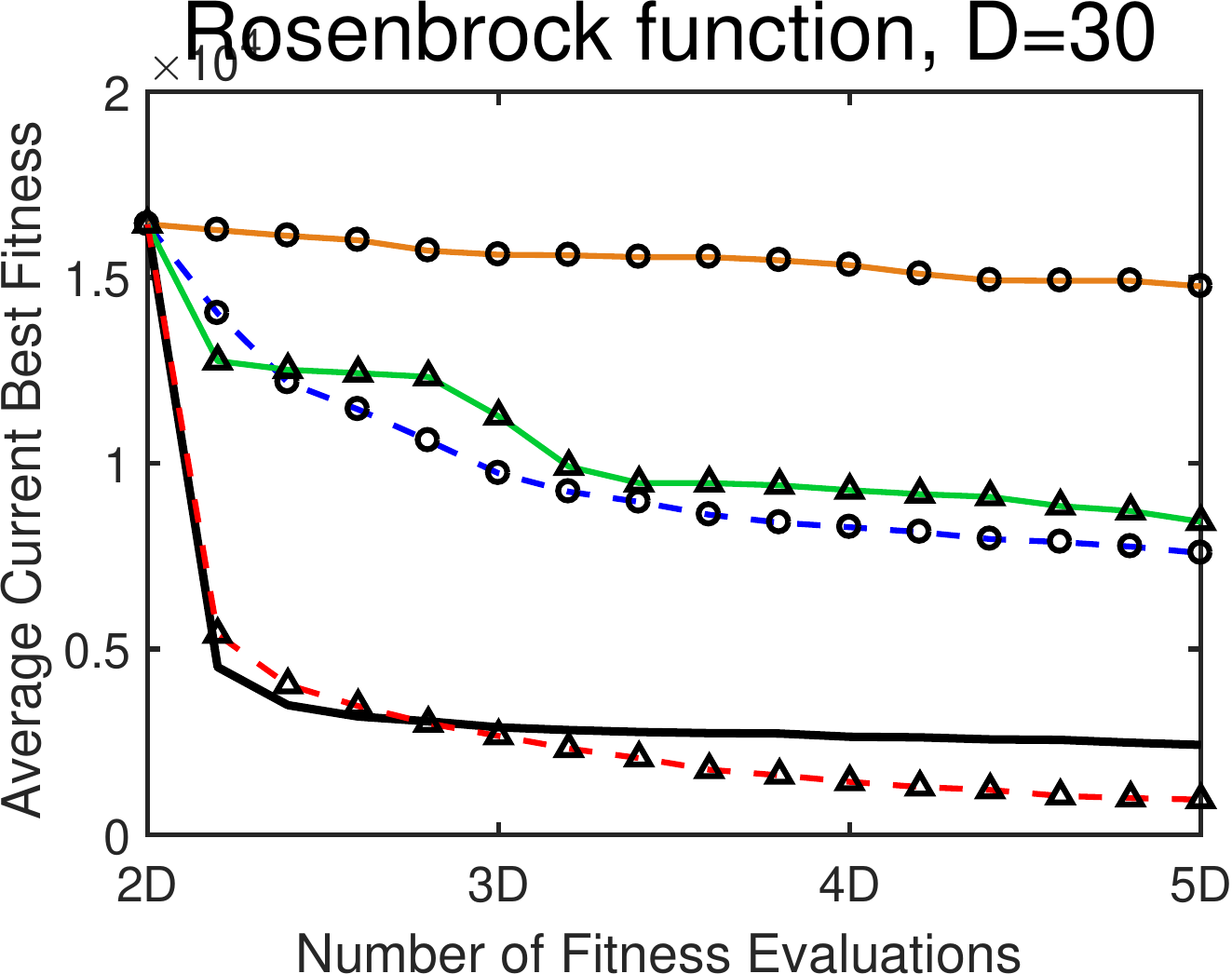}\quad 
    \includegraphics[width=.18\linewidth]{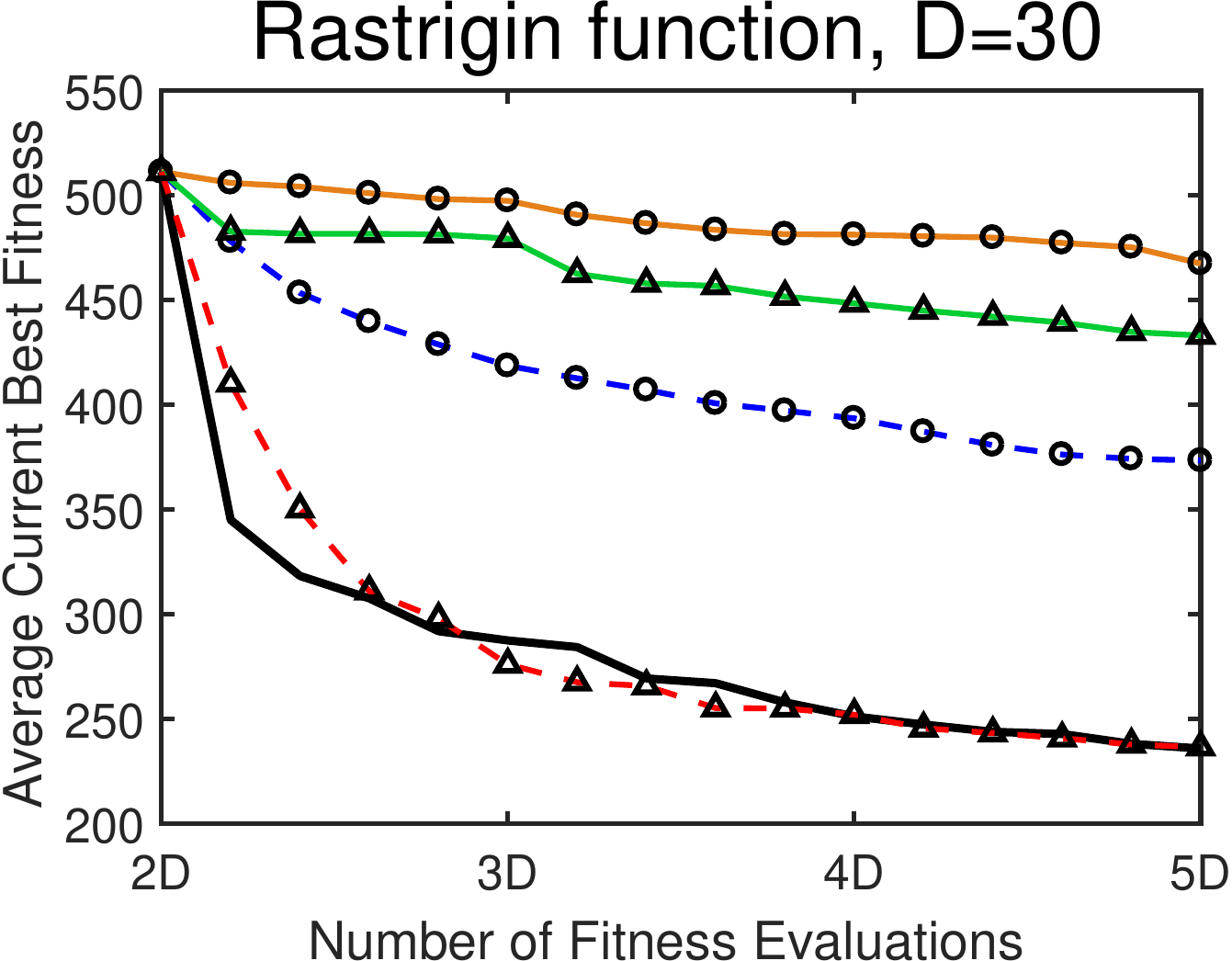}
    \medskip

    \includegraphics[width=0.4\linewidth]{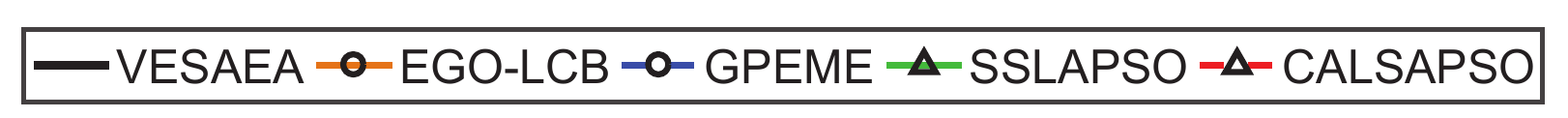}
    
    \caption{Convergence curves of VESAEA, EGO-LCB, GPEME, SSLAPSO and CALSAPSO on benchmark problems with $D=5,15,30$. }
    \label{sphere-griewank}
    \end{figure*}

The results obtained by five algorithms over 25 independent runs are shown in Table \ref{result} and the convergence curves of algorithms for all test problems are plotted in Fig. 
\ref{sphere-griewank}, where the x-axis ranges from $2D$ to $5D$ because the first $2D$ fitness evaluations are used for initialization, which is same for all algorithms in each run for fair comparison. Due to the page limitation, only results of $D=5, 15, 30$ are shown in the paper.  The average rank is calculated according to the best fitness on each test problem. The Friedman statistical test and its postdoc test, Nemenyi test \cite{derrac2011practical}, are both carried out with a 0.05 significance level. The $p$-value by Friedman test equal to $6.193e-12$ lower than $0.05$ significantly, revealing significant difference among five algorithms. Then the Nemenyi test result is presented as adjust $p$-value in Table \ref{result} where VESAEA is the control method. It also shows the performance of VESAEA is greatly different with EGO-LCB, GPEME, SSLAPSO and CALSAPSO. In summary, considering average rank and statistical testing results, VESAEA significantly outperforms the other four algorithms. 

From Table \ref{result} and the convergence curves, we can find the performance of VESAEA is similar in sphere, griewank and ackley functions, in which VESAEA obtains the best results both in convergence behaviour and the quality of final solution for $D=15,20,30$. Meanwhile, the EGO-LCB is slightly more efficient than others when the dimension $D=5$ in these three functions. Actually, the sphere problem is an absolutely uni-modal problem while the griewank and ackley problem can be regarded as relatively uni-modal problem except for many small peaks in the whole landscape for griewank problem and a narrow hole in the centre of ackley function. Therefore, we can conclude the proposed model management strategy is efficient in the overall uni-modal whether with or without small peaks.  

% The performance of the proposed framework is also very similar in these two functions whose evolution curves are presented in Fig.~\ref{sphere-griewank}. Compared with the other four algorithms, VESAEA obtains the best results both in convergence behaviour and the quality of final solution for $D=15,30$. By contrast, the EGO-LCB is slightly more efficient than others when the dimension $D=5$. As a sequence, we could generally conclude the proposed model management strategy is efficient in the \hao{overall uni-modal} whether with or without small peaks. 

% For Ackley function, which has a nearly flat outer region and a large hole at the centre, the VESAEA is slightly better than other four algorithms for $D=15, 30$ while the EGO-LCB performs best for $D=5,10$ as shown in the first column of Fig. \ref{ackley-rosenbrock}. Also for $D=15, 30$, we could easily observe that CALSAPSO performs well at the beginning of search process but is surpassed by VESAEA in the latter stage of the search, which indicates that the new framework is capable to balance better the exploitation and exploration. 

Different with the above problems, Rosenbrock function is a multi-modal problem with a very narrow valley. VESAEA obtains the best results for problems of $D=5, 10, 15$ and the convergence profile is presented in the fourth column of Fig. \ref{sphere-griewank}. From evolution processes, it is obvious that VESAEA outperforms other four methods totally for problems of $D=15$. And for the lowest dimension case $D=5$, VESAEA is also not very efficient in the early search stage but still gets best result finally similar to former cases in Ackley problems. However for $D=30$, the CALSAPSO converges slightly better than VESAEA in the latter stage of the search and obtains higher-quality results. 

For the most complicated test problem, Rastrigin function, which has many significant peaks in the landscape, the performance of VESAEA framework is better than EGO-LCB, GPEME and SSLAPSO and similar to CALSAPSO of as shown in the last column of Fig. \ref{sphere-griewank}. It is probably because all five algorithms are hard to jump out of the local optimal that have strong capacity of trapping the optimization algorithm within limited fitness evaluations. 

From the above observations, we can see that the VESAEA is capable of solving the uni-modal problems even though with little noise in the search space. VESAEA obtains the best result in 8 out of 10 test cases which indicates its capability in optimizing low-dimension of problems. But the classical framework, EGO-LCB, is always the best algorithm in very low-dimension problems, like $D=5$. Besides, experimental results also indicate that it is hard for GPEME and SSLAPSO to find a high quality solution in such limited computational resources. 
% Although it is hard to build much accurate surrogate in very limited evaluated samples, VESAEA trades off the global search and local search effectively to optimize the expensive problems.

However, the performance of VESAEA deteriorates in very complicated problems with many attractive local optimum, like Rastrigin problem. For this kind of problems, the global search hardly jumps out of the local optimums with limited fitness evaluations and on the other hand, the Voronoi based local search only focuses on the better area in the current generation. As a sequence, the algorithm is easily trapped in the local optimal especially for high dimension cases.

\begin{figure}[htpb!]
    \centering
    \includegraphics[width=0.8\linewidth]{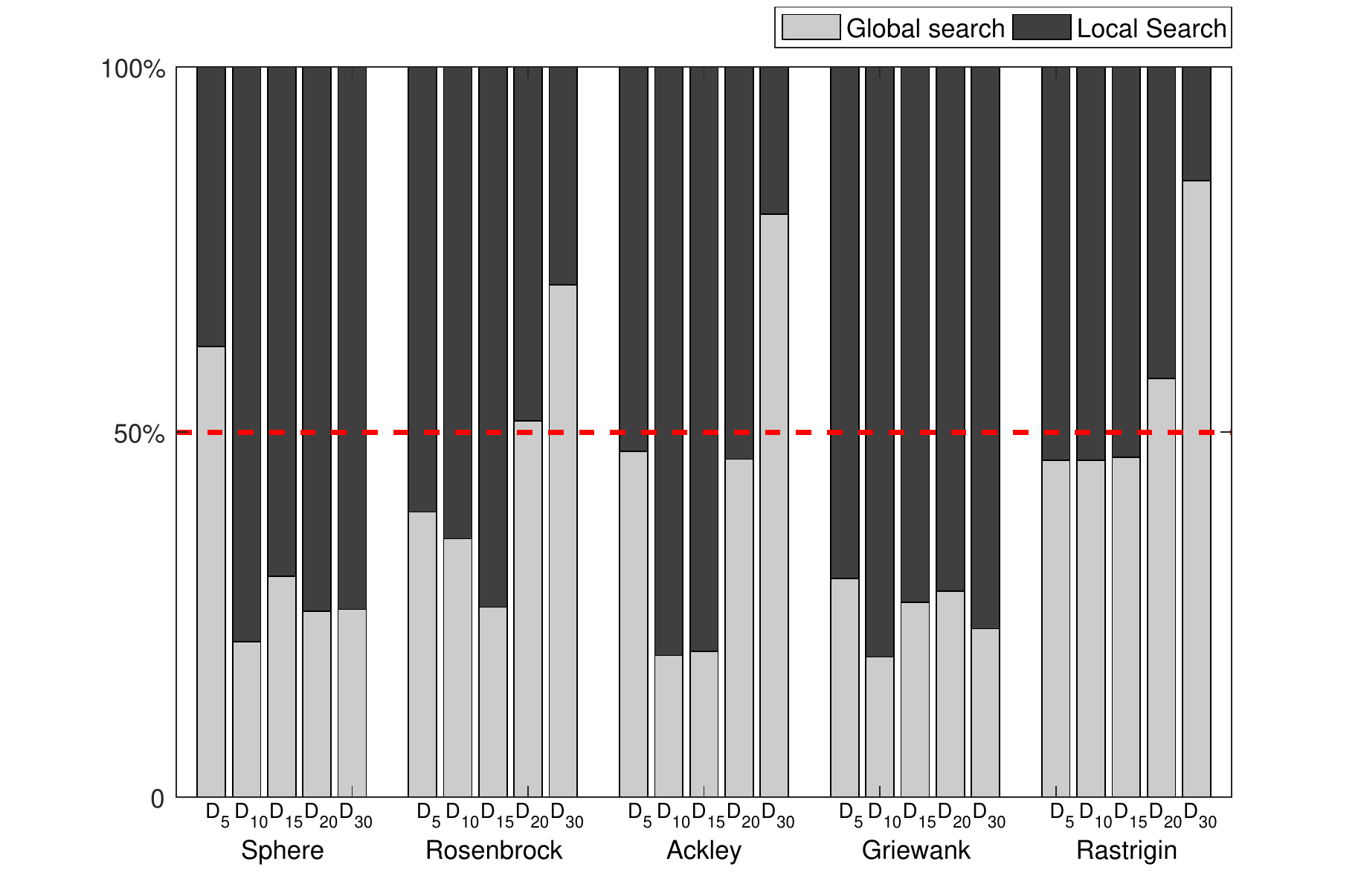}
    \caption{Average contributing ratio of each search stage over 25 trials.}
    \label{ratio}
\end{figure}

\begin{figure}[htpb!]
    \centering
    \includegraphics[width=.4\linewidth]{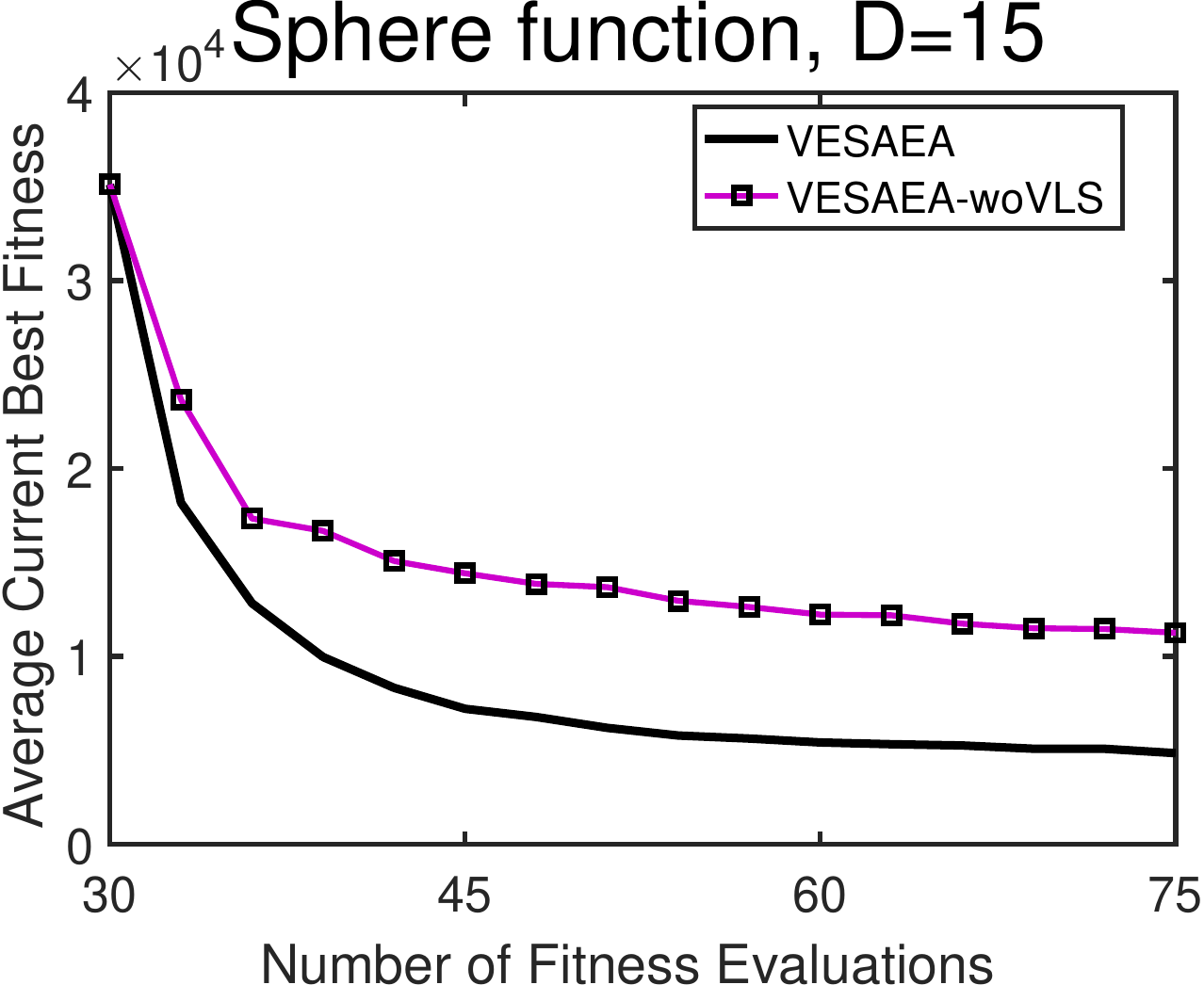}%\quad
    \includegraphics[width=.4\linewidth]{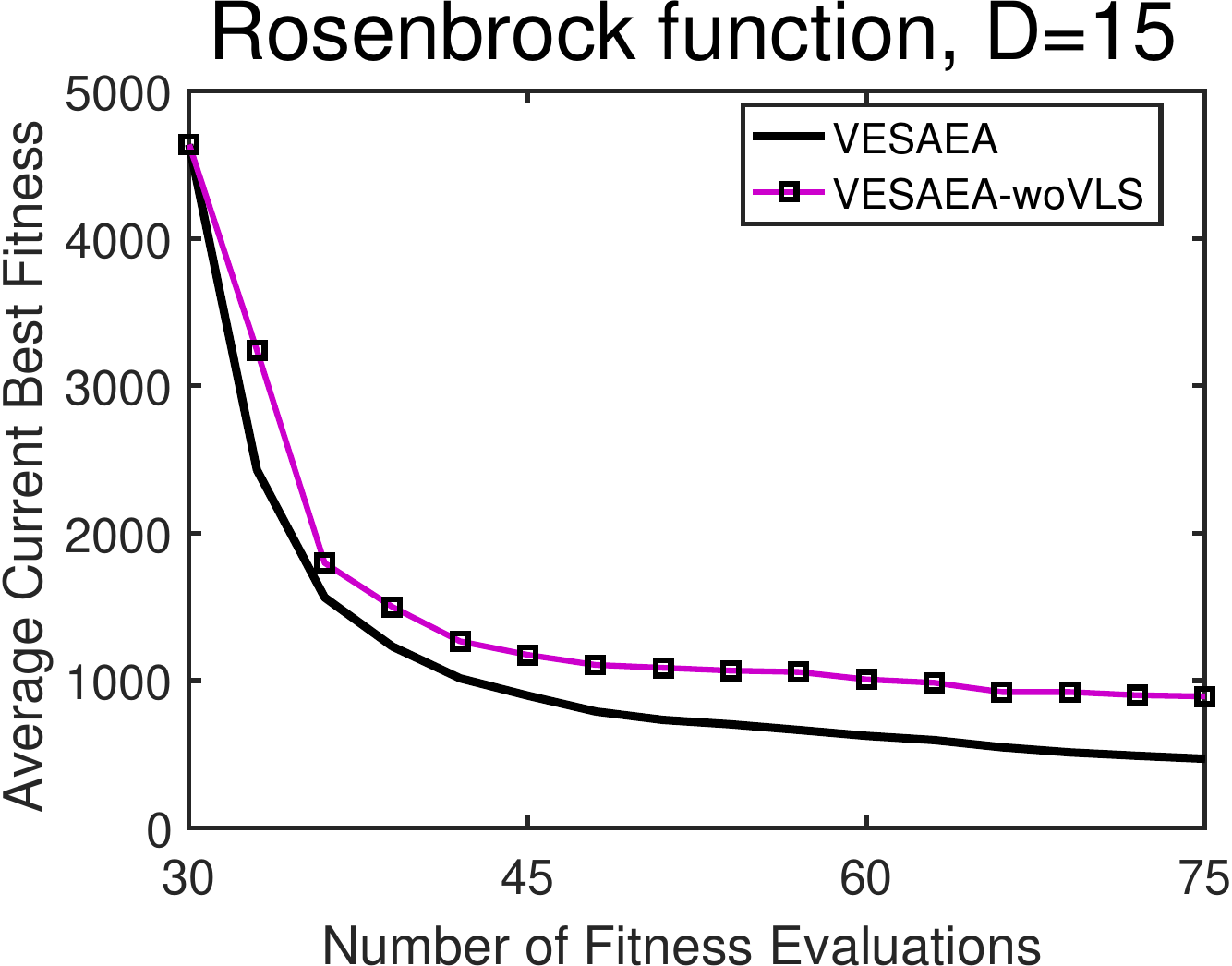}%\quad
    \medskip
    
    \includegraphics[width=.4\linewidth]{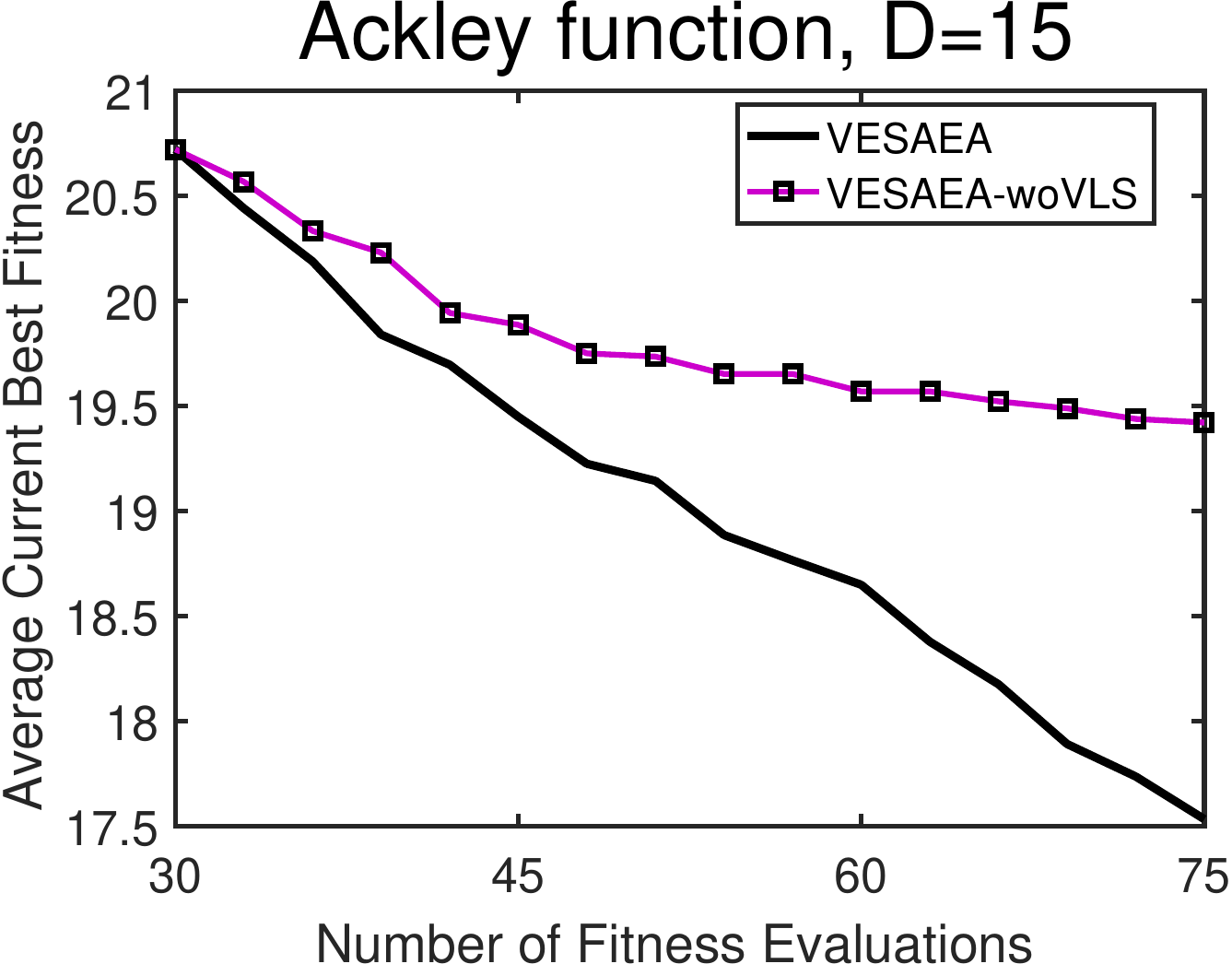}%\quad
    \includegraphics[width=.4\linewidth]{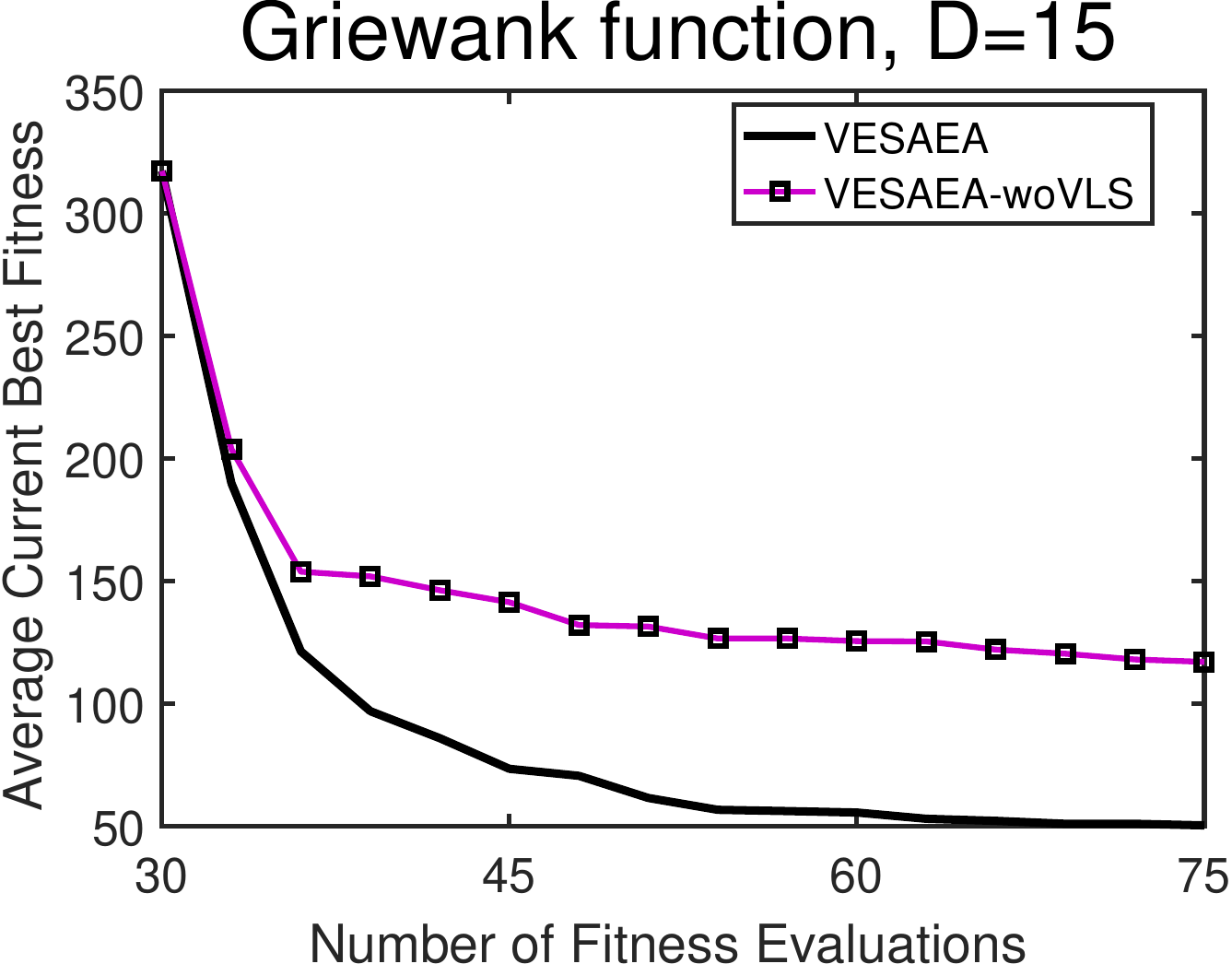}%\quad
    \medskip
    
    \includegraphics[width=.4\linewidth]{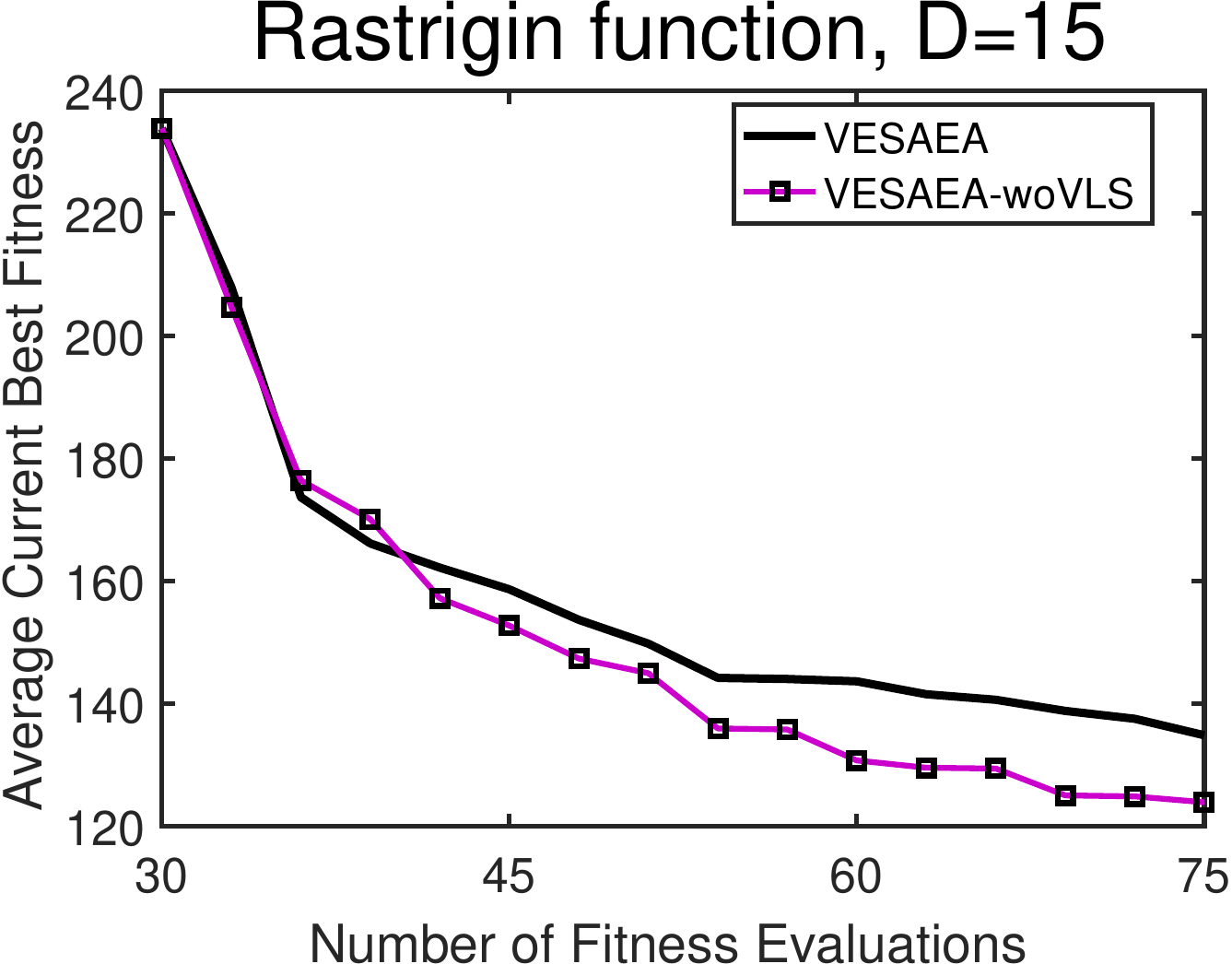}%\quad

    \caption{Convergence curves of VESAEA and VESAEA-woVLS on five problems with $D=15$.}
    \label{VESAEA-woVLS}
    \end{figure}

\subsection{The effect of Voronoi-based local search}

In this paper, we creatively employed Voronoi diagram for local search in SAEA framework. In order to study the effect of Voronoi based local search furtherly, we analyzed the average contribution ratio of local search and global search stages over all 25 independent runs. The results are presented in Fig. \ref{ratio} where the red dotted line represents bisector of contribution ratio. 

In Fig. \ref{ratio}, there are 19 out of 25 test instances in which the local search contributes more than the global search. We can find in griewank and sphere problems, the Voronoi-based local search accounts for a large proportion (over than 70\%) of the whole search process, while in rastrigin and rosenbrock problems, two search stages make about the same contributions to the final result. Moreover, we calculate the relation between dominated search stage and the optimization result as presented in Table \ref{local-global}. From the table, we can find local search dominated VESAEA accounts for 14 cases and the global search dominated VESAEA only has 3 cases among 17 test instances in which VESAEA obtains the best result. 
\begin{table}[htpb!]
    \centering
    \caption{Relation between dominated search stage and the obtained result: the number that local search dominated or global search dominated VESAEA obtains the best or not best result. }\label{local-global}
    \begin{tabular}{|c|c|c|}
    \hline
    \multirow{2}{*}{}      & \multicolumn{2}{c|}{Dominated Search} \\ \cline{2-3} 
                           & Local Search      & Global Search     \\ \hline
    VESAEA is the best     & 14                  & 3                 \\ \hline
    VESAEA is not the best & 5                  & 3                 \\ \hline
    \end{tabular}
    \end{table}

\begin{table}[htpb!]
    \caption{Averaged best fitness and standard deviation obtained by VESAEA and VESAEA-woVLS on five test problems with $D=15$ over 25 independent runs. The boldface figures are the best fitness among five algorithms in each test problem.}\label{local-global-result}
    \begin{tabular}{|c|c|c|c|}
    \hline
    Problem    & D  & VESAEA                          & VESAEA-woVLS                    \\ \hline
    Sphere     & 15 & \textbf{4.85e+03 $\pm$ 7.64e+02} & 1.13e+04 $\pm$ 1.74e+03          \\ \hline
    Rosenbrock & 15 & \textbf{4.69e+02 $\pm$ 1.74e+02} & 8.93e+02 $\pm$ 2.56e+02          \\ \hline
    Ackley     & 15 & \textbf{1.75e+01 $\pm$ 1.19e+00} & 1.94e+01 $\pm$ 4.67e-01          \\ \hline
    Griewank   & 15 & \textbf{5.04e+01 $\pm$ 1.04e+01} & 1.17e+02 $\pm$ 1.50e+01          \\ \hline
    Rastrigin  & 15 & 1.35e+02 $\pm$ 2.21e+01          & \textbf{1.24e+02 $\pm$ 2.23e+01} \\ \hline
    \end{tabular}
    \end{table}

To further assess the capability of Voronoi-based local search, we compare VESAEA and VESAEA without Voronoi-based local search stage (VESAEA-woVLS) on 5 test problems of dimension $D=15$ with 25 independent runs. The results are presented in Table \ref{local-global-result} and the convergence curves are presented in Fig. \ref{VESAEA-woVLS} in which the setting is same as the previous experiments. It is clear that VESAEA performs significantly better than VESAEA-woVLS in sphere, ackley and griewank problems, probably because the better Voronoi cells usually cover the area containing global optimum instead of local optimum. And it has a similar performance to VESAEA-woVLS in rosenbrock problems because the best Voronoi cells might all fall into the narrow valley and VESAEA can not benefit much from the Voronoi-based local search. But in rastrigin problem, VESAEA-woVLS performs slightly better than VESAEA, probably because VESAEA-woVLS only focuses on the global search and easily jumps out of many attractive local minimums. In addition, we could find test problems in which VESAEA performs better than VESAEA-woVLS are consistent with problems in Fig. \ref{ratio} in which the local search contributes more than the global search. 

Overall, we can conclude that the Voronoi-based local search is able to improve the convergence of optimization, especially for uni-modal problems and multi-modal problems only with many small peaks. And the efficacy of Voronoi-based local search deteriorates in complicated multi-modal problems with many attractive local optimums.

\section{Conclusion}\label{conclusion}
A Voronoi-based efficient surrogate assisted evolutionary algorithm framework is proposed in this paper for very expensive problems with extremely limited computational resources. The framework mainly has two search stages including leave-one-out cross validation based global search and Voronoi based local search. The global search stage employs LOOCV and RBF model to improve the accuracy of surrogate model and explore the search space. In local search stage, the Voronoi diagram, which divides the whole space into many cells, is applied to assist the exploitation in local area. We designed a performance selector which switches between two search stages by detecting whether current search stage has improvements or not to trade off the exploitation and exploration. The efficacy of proposed framework is examined by comparing with a few state-of-the-art SAEAs on some commonly used test problems. The results demonstrate that the proposed framework is powerful at solving uni-modal problems, even though there are several small peaks around the response surface. Moreover, we evaluates the effect of Voronoi-based local search and the results show that Voronoi diagram contributes a lot in finding a better solution on a very limited computational budget.

The ability of Voronoi for boosting optimization in expensive problems has been demonstrated in this work. In the future, the efficacy of Voronoi-based optimization algorithms will be tested on more complicated test functions. On the other hand, more strategies with Voronoi assisted SAEA for very expensive problems will be considered. Since Voronoi diagram divides the whole search space into many cells, it is highly desirable to investigate more on Voronoi diagram for large-scale problems.

\section*{Acknowledgement}
This work was supported by National Key R$\&$D Program of China(Grant No. 2017YFC0804003), the Program for Guangdong Introducing Innovative and Enterpreneurial Teams (Grant No. 2017ZT07X386), Shenzhen Peacock Plan (Grant No. KQTD2016112514355531), the Science and Technology Innovation Committee Foundation of Shenzhen (Grant No. ZDSYS201703031748284, JCYJ20180504165652917, JCYJ20170307105521943 and JCYJ20170817112421757) and the Program for University Key Laboratory of Guangdong Province (Grant No. 2017KSYS008).

\bibliographystyle{IEEEtran}
\bibliography{reference}
\end{document}